\PassOptionsToPackage{usenames,dvipsnames}{xcolor}
\pdfoutput=1
\documentclass[11pt]{article}

\usepackage[final]{acl}
\usepackage{times}
\usepackage{latexsym}

\usepackage[T1]{fontenc}
\usepackage[utf8]{inputenc}

\usepackage{microtype}

\usepackage{inconsolata}

\usepackage{graphicx}

\usepackage{tcolorbox}\tcbuselibrary{skins}
\usepackage{xcolor}
\usepackage{colortbl}
\usepackage{csquotes}
\usepackage{booktabs}
\usepackage{amssymb}
\usepackage{amsmath} 
\usepackage[sans]{dsfont} 
\usepackage{tikz}
\usepackage{pifont}
\usepackage{multirow}
\usepackage{soul, color}
\usepackage{tcolorbox}
\usepackage{tabularx}
\usepackage{xspace}
\usepackage{makecell}

\definecolor{color1}{HTML}{EAFFF1}
\definecolor{color2}{HTML}{C9F1E4}
\definecolor{color3}{HTML}{A7E3D7}
\definecolor{color4}{HTML}{85D5CA}
\definecolor{color5}{HTML}{6CC5C0}
\definecolor{color6}{HTML}{58B4B9}
\definecolor{color7}{HTML}{44A2B1}
\definecolor{color8}{HTML}{3091AA}
\definecolor{color9}{HTML}{2880A0}
\newcommand*{\opacity}{50}% Here you can change the opacity of the background color!
\newcommand*{\minval}{0.05}% Define the minimum value on your data set
\newcommand*{\maxval}{0.9}% Define the maximum value in your data set!

\newcommand{\gradient}[1]{
    \ifdimcomp{#1pt}{>}{\maxval pt}{#1}{
        \ifdimcomp{#1pt}{<}{\minval pt}{#1}{
            \pgfmathparse{int(round(8*(#1/(\maxval-\minval))-(\minval*(8/(\maxval-\minval)))))}
            \xdef\tempa{\pgfmathresult}
            \ifcase\tempa
                \cellcolor{color1!\opacity} #1\or
                \cellcolor{color2!\opacity} #1\or
                \cellcolor{color3!\opacity} #1\or
                \cellcolor{color4!\opacity} #1\or
                \cellcolor{color5!\opacity} #1\or
                \cellcolor{color6!\opacity} #1\or
                \cellcolor{color7!\opacity} #1\or
                \cellcolor{color8!\opacity} #1\or
                \cellcolor{color9!\opacity} #1
            \fi
    }}
}

\newcommand{\fon}[1]{\fontfamily{#1}\selectfont} 
\definecolor{CB_lightCyan}{HTML}{99DDFF}
\definecolor{CB_pear}{HTML}{BBCC33}
\definecolor{CB_pink}{HTML}{FFAABB}
\newcolumntype{P}[1]{>{\centering\arraybackslash}p{#1}}
\newcommand{\llama}{\raisebox{-1.0pt}{\includegraphics[height=0.8em]{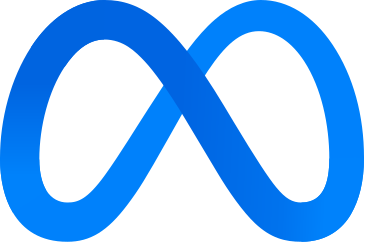}}\xspace}
\newcommand{\gpt}{\raisebox{-1.0pt}{\includegraphics[height=0.9em]{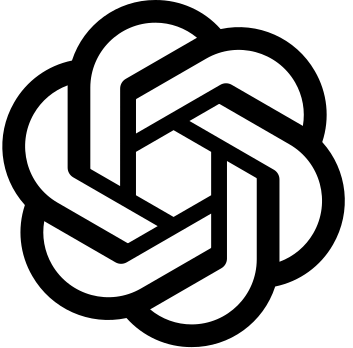}}\xspace}
\newcommand{\gemma}{\raisebox{-1.5pt}{\includegraphics[height=1em]{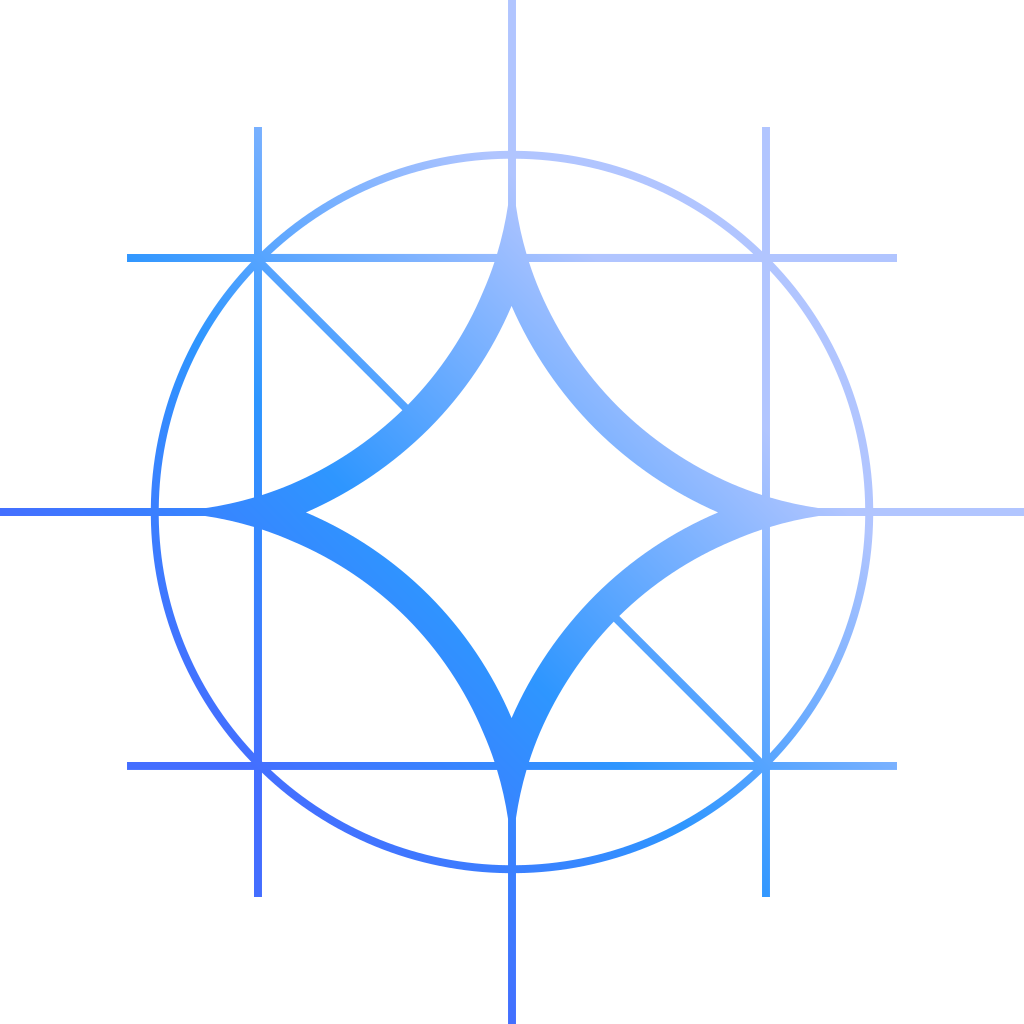}}\xspace}
\newcommand{\phim}{\raisebox{-1.5pt}{\includegraphics[height=0.9em]{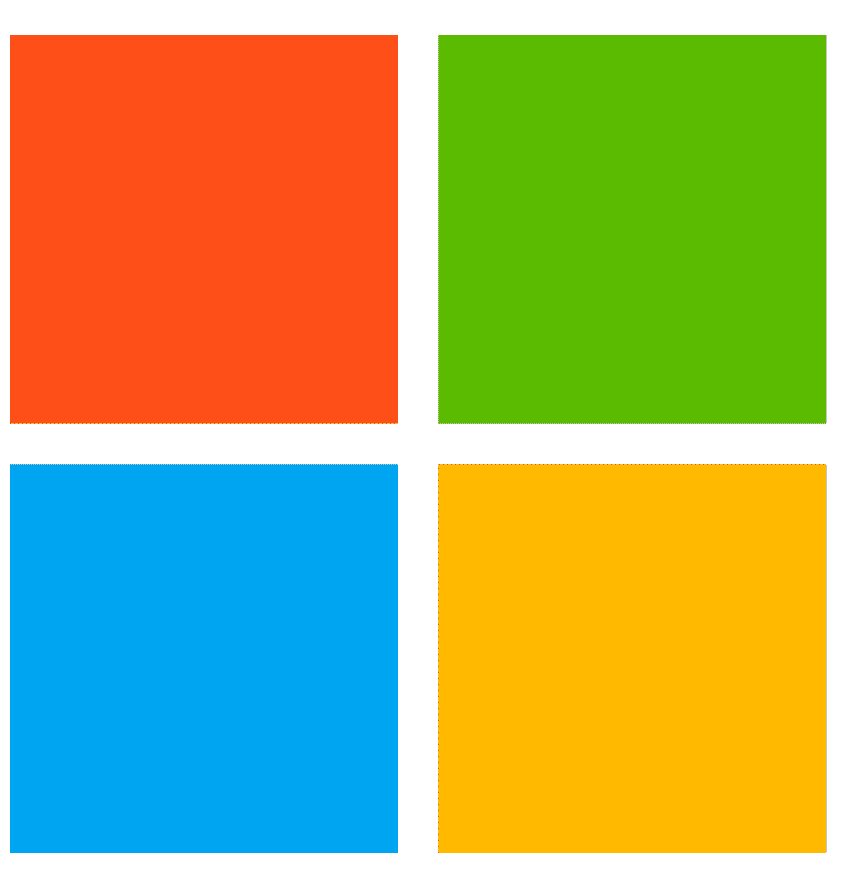}}\xspace}

\newcommand{\bluecircle}{\raisebox{-1.5pt}{\includegraphics[height=0.9em]{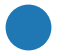}}\xspace}
\newcommand{\orangetriangle}{\raisebox{-1.5pt}{\includegraphics[height=0.9em]{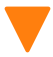}}\xspace}
\newcommand{\greensquare}{\raisebox{-1.5pt}{\includegraphics[height=0.9em]{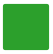}}\xspace}

\tcbset{prompt/.style={
    enhanced,
    size=fbox,
    boxrule=2pt,
    arc=2mm,
    auto outer arc,
    left=1pt,
    right=1pt,
    top=1pt,
    bottom=1pt,
    fontupper=\fon{cmtt}, 
    colback=CB_lightCyan!15,
    colframe=CB_lightCyan!30,
    coltitle=CB_lightCyan!25!black, 
}}

\definecolor{vanilla}{RGB}{255, 241, 218}
\definecolor{lightgreen}{RGB}{230, 255, 239}
\definecolor{lightblue}{RGB}{253, 198, 212}
\definecolor{lightred}{RGB}{202, 185, 253}
\DeclareRobustCommand{\hlcolor}[1]{{\sethlcolor{vanilla}\hl{#1}}}

\title{Tell, Don't Show: Leveraging Language Models'\\ Abstractive Retellings to Model Literary Themes}

\author{
\textbf{Li Lucy}$^1$ \quad
\textbf{Camilla Griffiths}$^2$ \vspace{0.2em} \\
\textbf{Sarah Levine}$^2$ \quad
\textbf{Jennifer L. Eberhardt}$^2$ \quad
\textbf{Dorottya Demszky}$^2$ \vspace{0.2em} \quad 
\textbf{David Bamman}$^1$ \vspace{0.3em}
\\
$^1$University of California Berkeley \quad
$^2$Stanford University 
\vspace{0.3em} \\
\texttt{lucy3\_li@berkeley.edu} 
  }

\begin{document}
\maketitle
\begin{abstract}
Conventional bag-of-words approaches for topic modeling, like latent Dirichlet allocation (LDA), struggle with literary text. Literature challenges lexical methods because narrative language focuses on immersive sensory details instead of abstractive description or exposition: writers are advised to \textit{show, don't tell}. We propose \texttt{Retell}, a simple, accessible topic modeling approach for literature. Here, we prompt resource-efficient, generative language models (LMs) to \textit{tell} what passages \textit{show}, thereby translating narratives' surface forms into higher-level concepts and themes. By running LDA on LMs' retellings of passages, we can obtain more precise and informative topics than by running LDA alone or by directly asking LMs to list topics. To investigate the potential of our method for cultural analytics, we compare our method's outputs to expert-guided annotations in a case study on racial/cultural identity in high school English language arts books. 
\end{abstract}

\section{Introduction} \label{sec:intro}
A common task in text analysis is the identification and exploration of latent themes or topics across documents, and unsupervised topic modeling has sustained its popularity in cultural analytics \cite{boyd2017applications}. Latent Dirichlet allocation (LDA) remains popular \cite{blei2003latent, luhmann2022digital, fontanella2024we, sobchuk2024computational}, though the strength of language models (LMs) have raised new possibilities \citep[e.g.][]{pham-etal-2024-topicgpt,wan2024tnt}. While the rise of high-performing LMs in natural language processing (NLP) has opened new avenues for cultural analytics research, it has also raised barriers for access, especially since researchers in the humanities may be constrained by API costs and compute resources. 

Given this interdisciplinary context, we examine how one may perform topic modeling on literary passages with ``small'', resource-efficient LMs. We use these LMs to address a key challenge of computational literary analysis: literature often contains low-level depictions of characters' actions, thoughts, and dialogue, and so higher-level themes may evade classic text analysis tools reliant on lexical surface forms. This challenge of identifying patterns across documents is especially conspicuous for literary text: the golden rule of narrative writing advises authors to \textit{show, don't tell} \cite{lubbock1921craft, burroway2019writing}. That is, good storytelling should allow readers to experience a narrative through low-level sensory detail instead of high-level abstractive exposition. 

\begin{figure}[t]
    \includegraphics[width=\columnwidth]{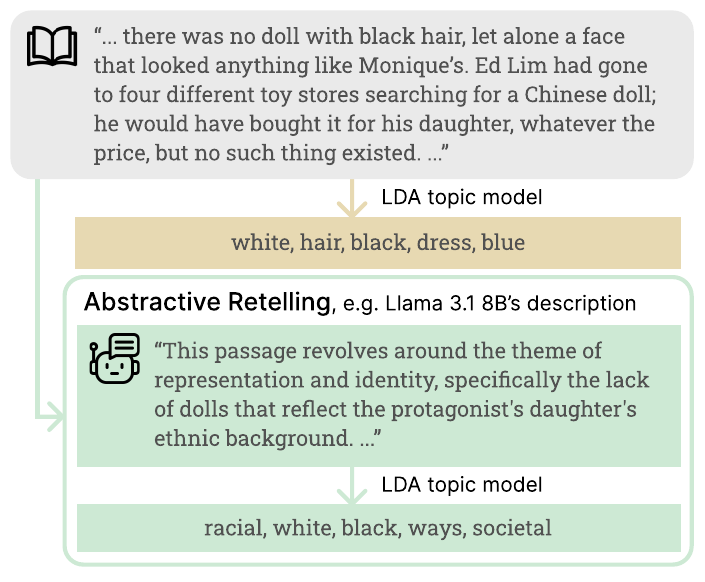}
    \centering
    \caption{Language models (LMs) can support traditional topic modeling pipelines, e.g. LDA, by translating low-level sensory detail (\textit{showing}) into high-level abstractive exposition (\textit{telling}). This example book excerpt is from \textit{Little Fires Everywhere} by Celeste Ng.}
	\label{fig:fig_1}
\end{figure}

To address this issue, we ask LMs to \textit{tell} us what book passages \textit{show} to better surface topics and themes shared across them. We propose a novel yet simple approach for running topic modeling on literary texts: instead of applying LDA on the original texts, we apply it on LMs' abstractive retellings of them (Figure~\ref{fig:fig_1}). We show that the topics we then obtain are more useful for identifying cross-cutting themes than topics outputted by LDA alone and than those obtained by directly requesting topic labels from the same LMs (\S\ref{sec:results}). 

Aside from evaluating our approach across a variety of topics, we closely examine its potential for supporting cultural analytics research in a case study. We apply our method on books taught in American high schools, and work with in-domain experts to identify passages that mention and/or discuss racial/cultural identity. Our approach is able to predict passages' topic distributions in a manner that corresponds with human annotators' codes, and identifies more topics and passages of interest than baseline methods. Overall, our conceptual framework of \textit{abstractive retelling} holds promise for literary content analysis. We release our code at \url{https://github.com/lucy3/tell_dont_show}.

\section{Related Work}

Topic modeling allows practitioners to model large collections of text in a bottom-up, discovery-oriented manner. Many topic modeling approaches have been proposed over the years, including structural topic models \cite{roberts2013structural}, hierarchical models \cite{griffiths2003hierarchical}, semi-supervised models \cite{jagarlamudi-etal-2012-incorporating, gallagher-etal-2017-anchored}, and embedding-based or neural models \cite{dieng2020topic, card2018neural, bianchi-etal-2021-pre, burkhardt2019decoupling, sia-etal-2020-tired}. Still, \citet{blei2003latent}'s LDA, a generative probabilistic model, remains popular as an analytical tool \citep[e.g.][]{zhu2024multimodal}, given its well-studied output behavior, stability, and ease of use \cite{hoyle-etal-2022-neural, mccallum2002mallet}. 

Recently, approaches that directly elicit topic labels from LMs have emerged \cite{pham-etal-2024-topicgpt, wan2024tnt, lam2024concept}, though these frameworks often involve an intricate chain of prompting steps. In contrast, our proposed improvement over traditional topic modeling requires only one prompt. Our method is most similar to prior work asking LMs to describe documents in a manner that surfaces information that is \textit{implicit} in them \cite{zhong2022describing, hoyle-etal-2023-natural, hua-etal-2024-get, ravfogel2024descriptionbased}. 

Given our focus on literature, our work also contributes to computational humanities \cite{piper2018enumerations, underwood2019}. At this junction, computational text analysis supports ``distant reading'' \cite{moretti2013distant, janicke2015close}, though some recent work has also examined the use of LMs for close, interpretative reading \cite{sui2025kristevaclosereadingnovel}. Topic models have been applied to a range of textual forms, including books \cite{jockers2013significant, thompson-mimno-2018-authorless, Erlin2017Topic}, historical newspapers \cite{newman2006, yang-etal-2011-topic}, folklore \cite{tangherlini2013trawling}, and poetry \cite{navarro2018poetic, rhody2topic}. Topic modeling's appeal in cultural analytics stems from its potential to identify content of interest and relate texts to each other \cite{tangherlini2013trawling}. Our evaluation approach and case study also consider these use cases (\S\ref{sec:results}, \S\ref{sec:case_study}). 

\section{Methods}\label{sec:methods}

Our task is to inductively discover latent topics across literary passages. We experiment with instruction-tuned LMs designed to be resource-efficient within various model families: GPT-4o mini, Llama 3.1 8B, Phi-3.5-mini (3.8B), and Gemma 2 2B. 

\subsection{Our Approach: \texttt{Retell}}\label{sec:our_method}

Our proposed approach for running topic modeling on literary passages first asks an LM to ``retell'' the contents of a literary passage. We define a passage to be a sequence of paragraphs from a book, containing up to 250 white-spaced tokens in total. We experiment with three retelling \texttt{VERB}s -- \textit{describe}, \textit{summarize}, or \textit{paraphrase} -- in a short prompt: 

\begin{displayquote}
In one paragraph, \texttt{[VERB]} the following book excerpt for a literary scholar analyzing narrative content. Do not include the book title or author's name in your response; \texttt{[VERB]} only the passage. \\ \\ Passage: \\ \texttt{[PASSAGE]}
\end{displayquote}

The first two verbs, \textit{describe} and \textit{summarize}, encourage abstraction, while \textit{paraphrase} does not. Since abstraction differentiates \textit{telling} from \textit{showing}, experimenting with several verbs allows us to see if more abstractive variants of \texttt{Retell} yield better results. We run each LM with default parameters (Appendix~\ref{sec:appdx_params}). 

We run latent Dirichlet allocation (LDA) using Mallet\footnote{Release version \texttt{202108}.} \cite{mccallum2002mallet} on models' passage retellings. For LDA, we lowercase all text and remove words with fewer than 3 characters, frequent words used by more than 25\% of all retellings, and rare words used by fewer than 5 retellings; these text preprocessing steps match those of \citet{thompson-mimno-2018-authorless}. We also remove characters' names detected by spaCy's \texttt{en\_core\_web\_trf} named entity recognition model. We prompt models to not mention book titles or author names and remove character names from outputs to avoid producing topics that only cluster terms associated with individual books \cite{jockers2013macroanalysis}. On average, LMs' retelling lengths range from Gemma 2 2B \texttt{Retell}-\textit{summarize}'s 105.5 words to GPT-4o mini \texttt{Retell}-\textit{describe}'s 170.0 words.\footnote{We use \texttt{blingfire}'s word tokenization to calculate these counts: \url{https://github.com/microsoft/BlingFire}.}

\subsection{Baselines}

\paragraph{Latent Dirichlet allocation.} Our first baseline runs LDA on the original passages. Like in \S\ref{sec:our_method}, we run the same vocabulary filtering steps and remove characters' names from passages. We compare \texttt{Retell} to default LDA with various numbers of topics $k$. 

\paragraph{TopicGPT-lite.} Our second baseline is based on \citet{pham-etal-2024-topicgpt}'s TopicGPT framework, which directly elicits topic labels from an LM in a two-stage process. First, during topic generation, an LM suggests topics over a sampled subset of $N$ documents, accumulating a set of possible labels. Second, during topic assignment, the LM assigns topics from the topic set to all documents. Unlike with LDA, $k$ is not preset, but instead refers to the number of unique topics assigned by the LM. In other words, $k$ is determined by the outcome of TopicGPT itself. 

\citet{pham-etal-2024-topicgpt} originally developed TopicGPT with GPT-4, and have suggested that it is impractical to use with weaker LMs, e.g. Mistral-7B-Instruct. That is, during topic generation, the list of topics accumulated by weaker LMs can snowball into a large collection of overly specific labels. We observe similar issues, where Llama 3.1 8B generates 486 unique topics for only 100 documents.
% leading to increasingly slow inference time, high memory usage, and the LM getting ``lost in the middle'' \cite{liu2024lost}. 

We modify TopicGPT to make it feasible to run with smaller LMs, and call our adapted version \textit{TopicGPT-lite}. TopicGPT-lite asks LMs to suggest only one topic, rather than multiple, per document during the topic generation stage. LMs are still allowed to output multiple topics during the assignment stage. Our modification mitigates the overly high $k$ issue: Llama 3.1 8B produces $k=206$ unique topics assigned to all passages, Phi-3.5-mini produces $k=538$, and Gemma 2 2B produces $k=157$. Interestingly, GPT-4o mini tends to under-generate topics rather than over-generate them like other LMs do. So we run two versions using this LM: one with multi-label topic generation, as originally intended by TopicGPT, and one with single-label generation, matching our setup for other LMs. With the former, GPT-4o mini produces $k=89$ topics and with the latter, $k=14$. In the main text, we present results for $k= 89$, as it turns out to perform better; results for $k=14$ can be found in Appendix~\ref{sec:appdx_eval}. 

Since smaller LMs struggle more with instruction following, we minimally edited TopicGPT's prompts for TopicGPT-lite to encourage simple output formatting, or one topic per line. On average, the number of topics assigned per passage ranges from 2.8 (GPT-4o mini) to 9.8 (Llama 3.1 8B). To make TopicGPT-lite's behavior more comparable to LDA's during evaluation, our topic assignment prompt asks LMs to output topics in order of their prominence in the document. Our prompts can be found in Appendix~\ref{sec:apdx_topicgpt}. 

Our aim when adapting TopicGPT for use with smaller LMs is to test how our abstractive retelling approach introduced in \S\ref{sec:our_method} compares to directly eliciting topics from an LM, \textit{given similar resource constraints and uses}. Though their costs are similar, TopicGPT-lite requires more compute and runtime. That is, \texttt{Retell} prompts an LM once per passage and then runs LDA over them, while TopicGPT-lite involves longer prompts per passage and also runs an additional generation prompt over $N$ passages.\footnote{In our experiments, we set $N=1000$. \citet{pham-etal-2024-topicgpt} recommends choosing an $N$ based on one's compute budget, and $N > 600$ was sufficient for the datasets in their paper.} Another practical benefit of \texttt{Retell} is that the granularity of topics ($k$) can be quickly adjusted without rerunning any LM. TopicGPT-lite, on the other hand, may require additional prompt engineering to steer $k$. 

\section{Evaluation Setup} \label{sec:eval}

We evaluate the general use of our proposed topic modeling method in two ways. First, we obtain pre-existing literary themes/tags from online resources, and use them to efficiently evaluate topic models' performance across passage sets spanning a wide range of topics (\S\ref{sec:eval_passage_sets}). Second, we conduct human evaluation of topics outputted at the passage-level, using a setup similar to a classic topic intrusion test (\S\ref{sec:passage_eval}). Overall, our evaluation setup in this section aims to broadly assess the interpretive utility of topic predictions. 

\subsection{Relatedness of Passage Sets' Labels}\label{sec:eval_passage_sets}

A useful topic model should recover latent topics that resemble gold ones. Here, we examine how well each topic modeling approach draws out latent topics across a set of passages. To create a dataset of literary passages, we draw English literature from both Project Gutenberg (pre-1925) and contemporary book lists (1923-2021), e.g. bestseller lists (Appendix~\ref{sec:apdx_data}). 
%Our contemporary dataset is similar to those found in prior work \cite{chang-etal-2023-speak, lucy-bamman-2021-gender}. 
In total, we begin with 50.7k unique titles, which eventually result in 256 books from Project Gutenberg and 476 contemporary books after filtering to those with passages matched to gold labels. 

We obtain gold topic labels using readers' tagged book quotes on Goodreads, a popular book discussion and reviewing platform, and theme-labeled quotes on SparkNotes and Litcharts, two online study guide publishers. Tags and themes on these sites may reference related or duplicate concepts, e.g. \textit{women and sexism} and \textit{sexism and women's roles}. Thus, we manually recode scraped tags/themes into 27 general topics, e.g. \texttt{gender}, some of which appear across multiple label sources. Our dataset is a multi-label one, e.g. a passage discussing interracial relationships would belong to both \texttt{love} and \texttt{race}. 

LDA requires sufficiently long input documents to yield robust topic estimates. We make the assumption that the preceding context of a quote does not deviate far from the tag/theme attached to the quote. We match quotes to book passages, where each passage includes the paragraph around the quote as well as preceding paragraphs for up to 250 white-spaced tokens. In total, we obtain 11.6k labeled passages (21.1k topic-passage pairs), with an average of 1.67 topics per passage. To encourage LDA to learn robust topics, we augment our set of labeled passages with a same-sized sample of random passages from our book collection. Our dataset of literary passages contains a total of 5.02M white-spaced tokens; Appendix~\ref{sec:apdx_data} provides additional details on dataset composition. 

\begin{table*}[t]
\centering
% \footnotesize
\def\arraystretch{0.8}
\setlength{\tabcolsep}{0.3em}
\resizebox{0.9\textwidth}{!}{
\begin{tabular}{l c | l | l}
\toprule
\textbf{Method} & $k$ & \multicolumn{1}{c}{SparkNotes \texttt{gender}} & \multicolumn{1}{|c}{Litcharts \texttt{time}} \\
\midrule
\texttt{Retell}-\textit{summ.} \llama{} & 50 & \hlcolor{\textit{societal, women, expectations, norms, men}} & \hlcolor{\textit{past, poignant, time, life, loss}}\\ 
\texttt{Retell}-\textit{desc.} \gpt{} & 50 & \hlcolor{\textit{women, gender, expectations, roles, norms}} &  
\textit{life, existential, death, existence, mortality} \\ 
Default LDA & 50 & \textit{n’t, got, say, man, know}  & 
\textit{n’t, know, just, your, get} \\ 
TopicGPT-lite \llama{} & 206 & \textit{family} & \textit{life}\\ 
TopicGPT-lite \gpt{} & 89 & \textit{loneliness} & \textit{loneliness}\\ 
\midrule
\textbf{Method} & $k$ & \multicolumn{1}{c}{Goodreads \texttt{social class}} & \multicolumn{1}{|c}{Litcharts \texttt{education}} \\
\midrule
\texttt{Retell}-\textit{summ.} \llama{} & 50 & \hlcolor{\textit{financial, economic, labor, business, work}}& 
\hlcolor{\textit{importance, education, knowledge, more, approach}}\\ 
\texttt{Retell}-\textit{desc.} \gpt{} & 50 & \hlcolor{\textit{economic, financial, wealth, labor, social}} &  
\hlcolor{\textit{knowledge, education, educational, intellectual, learning}} \\ 
Default LDA & 50 & \hlcolor{\textit{money, work, than, business, these}}  & 
\textit{n’t, know, just, your, get} \\ 
TopicGPT-lite \llama{} & 206 & \hlcolor{\textit{wealth}} & \textit{life}\\ 
TopicGPT-lite \gpt{} & 89 & \textit{injustice} & \textit{loneliness}\\ 
\bottomrule
\end{tabular}
}
\caption{Examples of the most highly ranked/probable topics for passage sets that share the same gold topic. In the left column, \llama{} = Llama 3.1 8B and \gpt{} = GPT-4o mini. \hlcolor{Highlighted} predictions are semantically related to gold labels. As a reminder, TopicGPT-lite self-determines $k$, while \texttt{Retell} and LDA allows us to preset $k$. On aggregate, abstractive \texttt{Retell}-based approaches make more related predictions than baseline methods (Table~\ref{tab:topic_relatedness}).}
\label{tab:example_topics}
\end{table*}

\begin{table*}[h]
% \footnotesize
\def\arraystretch{0.8}
\setlength{\tabcolsep}{0.3em}
\resizebox{\textwidth}{!}{
\begin{tabular}{c | ccccc | cccc | cc cc cc cc}
\toprule
 & \multicolumn{5}{c|}{Default LDA} & \multicolumn{4}{c|}{TopicGPT-lite} & \multicolumn{8}{c}{\texttt{Retell}-\textit{paraphrase}} \\ 
\cmidrule(lr){1-1} \cmidrule(lr){2-6} \cmidrule(lr){7-10} \cmidrule(lr){11-18} 

 & \multicolumn{5}{c|}{\textit{no language model}} & \gpt{} & \llama{} & \phim{} & \gemma{} & \multicolumn{2}{c}{\gpt{}} & \multicolumn{2}{c}{\llama{}} & \multicolumn{2}{c}{\phim{}} & \multicolumn{2}{c}{\gemma{}} \\ 
\cmidrule(lr){1-1} \cmidrule(lr){2-6} \cmidrule(lr){7-7} \cmidrule(lr){8-8} \cmidrule(lr){9-9} \cmidrule(lr){10-10} \cmidrule(lr){11-12} \cmidrule(lr){13-14} \cmidrule(lr){15-16} \cmidrule(lr){17-18}

Rating / $k$ & 89 & 206 & 538 & 157 & 50 
& 89 & 206 & 538 & 157
& 89 & 50 & 206 & 50 & 538 & 50 & 157 & 50 \\
\midrule

\ding{51} & \gradient{0.22} & \gradient{0.20} & \gradient{0.27} 
& \gradient{0.22} & \gradient{0.03} % default LDA
& \gradient{0.28} & \gradient{0.48} 
& \gradient{0.42} & \gradient{0.33} % topicGPT-lite
& \gradient{0.38} & \gradient{0.38} % paraphrase O
& \gradient{0.58} & \gradient{0.38} % paraphrase L
& \gradient{0.63} & \gradient{0.47} % paraphrase P
& \gradient{0.70} & \gradient{0.47} \\ % paraphrase G

\textbf{?} & \gradient{0.20} & \gradient{0.12} & \gradient{0.17} 
& \gradient{0.15} & \gradient{0.12} % default LDA
& \gradient{0.30} & \gradient{0.42} 
& \gradient{0.30} & \gradient{0.32} % topicGPT-lite
& \gradient{0.35} & \gradient{0.37} % paraphrase O
& \gradient{0.35} & \gradient{0.35} % paraphrase L
& \gradient{0.20} & \gradient{0.35} % paraphrase P
& \gradient{0.22} & \gradient{0.32} \\ % paraphrase G

\ding{55} & \gradient{0.58} & \gradient{0.68} & \gradient{0.57} 
& \gradient{0.63} & \gradient{0.85} % default LDA
& \gradient{0.42} & \gradient{0.10} 
& \gradient{0.28} & \gradient{0.35} % topicGPT-lite
& \gradient{0.27} & \gradient{0.25} % paraphrase O
& \gradient{0.07} & \gradient{0.27} % paraphrase L
& \gradient{0.17} & \gradient{0.18} % paraphrase P
& \gradient{0.08} & \gradient{0.22} \\ % paraphrase G
\bottomrule
\end{tabular}
}
\newline
\vspace*{1mm}
\newline
\resizebox{0.95\textwidth}{!}{
\begin{tabular}{c | cc cc cc cc | cc cc cc cc}
\toprule
& \multicolumn{8}{c|}{\texttt{Retell}-\textit{summarize}} & \multicolumn{8}{c}{\texttt{Retell}-\textit{describe}} \\ 
\cmidrule(lr){1-1} \cmidrule(lr){2-9} \cmidrule(lr){10-17} 

 & \multicolumn{2}{c}{\gpt{}} & \multicolumn{2}{c}{\llama{}} & \multicolumn{2}{c}{\phim{}} & \multicolumn{2}{c|}{\gemma{}} & \multicolumn{2}{c}{\gpt{}} & \multicolumn{2}{c}{\llama{}} & \multicolumn{2}{c}{\phim{}} & \multicolumn{2}{c}{\gemma{}} \\ 
\cmidrule(lr){1-1} \cmidrule(lr){2-3} \cmidrule(lr){4-5} \cmidrule(lr){6-7} \cmidrule(lr){8-9} \cmidrule(lr){10-11} \cmidrule(lr){12-13} \cmidrule(lr){14-15} \cmidrule(lr){16-17} 
Rating / $k$ & 89 & 50 & 206 & 50 & 538 & 50 & 157 & 50
& 89 & 50 & 206 & 50 & 538 & 50 & 157 & 50 \\
\midrule
\ding{51} & \gradient{0.65} & \gradient{0.60} % summarize O
& \gradient{0.67} & \gradient{0.58} % summarize L
& \gradient{0.62} & \gradient{0.50} % summarize P
& \gradient{0.62} & \gradient{0.50} % summarize G
& \gradient{0.67} & \gradient{0.68} % describe O
& \gradient{0.55} & \gradient{0.53} % describe L
& \gradient{0.63} & \gradient{0.63} % describe P
& \gradient{0.62} & \gradient{0.47} \\ % describe G 

\textbf{?} & \gradient{0.22} & \gradient{0.30} % summarize O
& \gradient{0.20} & \gradient{0.33} % summarize L
& \gradient{0.23} & \gradient{0.42} % summarize P
& \gradient{0.28} & \gradient{0.37} % summarize G
& \gradient{0.22} & \gradient{0.20} % describe O
& \gradient{0.35} & \gradient{0.40} % describe L
& \gradient{0.30} & \gradient{0.32} % describe P
& \gradient{0.23} & \gradient{0.42} \\ % describe G 

\ding{55} & \gradient{0.13} & \gradient{0.10} % summarize O
& \gradient{0.13} & \gradient{0.08} % summarize L
& \gradient{0.15} & \gradient{0.08} % summarize P
& \gradient{0.10} & \gradient{0.13} % summarize G
& \gradient{0.12} & \gradient{0.12} % describe O
& \gradient{0.10} & \gradient{0.07} % describe L
& \gradient{0.07} & \gradient{0.05} % describe P
& \gradient{0.15} & \gradient{0.12} \\ % describe G 
\bottomrule
\end{tabular}
}
\caption{Distributions of crowdworkers' ratings of relatedness between passage sets' most prominently predicted topic labels and their gold labels for GPT-4o mini (\gpt{}, $k=89, k = 50$), Llama 3.1 8B (\llama{}, $k=206, k = 50$), Gemma 2 2B (\gemma{}, $k=157, k = 50$), and Phi-3.5-mini (\phim{}, $k=538, k = 50$). On the left, \ding{51} = ``Very Related'', \textbf{?} = ``Somewhat Related'', and \ding{55} = ``Not Related''. Appendix~\ref{sec:appdx_topic_rel} includes additional results for GPT-4o mini and Llama 3.1 8B showing how \texttt{Retell}'s better ratings generalize with $k = 100, k = 200$.}
\label{tab:topic_relatedness}
\end{table*}

For each set of passages that share the same gold topic, we ask Prolific crowdworkers to rate on a scale of 1-3 how well the set's most prominently predicted topic relates to its gold one (Table~\ref{tab:example_topics}). For LDA-based approaches, the most prominent topic, e.g. \textit{school, class, college, boys, children}, is the one with the highest average probability. For TopicGPT-lite, the most prominent topic, e.g. \textit{childhood innocence}, is the one with the highest mean reciprocal rank \cite{voorhees2000trec}. A topic's reciprocal rank is $1/r_i$, where $r_i$ is the rank of topic $t$ for passage $i$. If the LM does not assign $t$ to $i$, then $1/r_i = 0$. 

In total, disaggregated across our three gold label sources, we evaluate the most prominently predicted topic of 60 passage sets, e.g. the set of passages labeled by Goodreads as \texttt{love}, or the set labeled by SparkNotes as \texttt{death}. We compare ratings provided for all three \texttt{Retell} variants (\textit{summarize}, \textit{describe}, \textit{paraphrase}) and our baselines (default LDA and TopicGPT-lite), with $k=50$ as well as for various $k$ determined by TopicGPT-lite. Crowdworkers achieve good inter-annotator agreement (weighted Cohen's $\kappa$ = 0.70). We pay workers \$16 an hour, and Appendix~\ref{sec:appdx_topic_rel} details worker recruitment, quality checks, and annotation instructions.

\subsection{Passage-level Topic Relevance} \label{sec:passage_eval}

We also evaluate our methods at the passage-level, using a setup inspired by \citet{chang2009reading}'s popular topic intrusion test. Here, we present annotators with a random shuffle of the top three topics predicted for a passage and one ``intruder'' topic. For LDA-based approaches, we sample the intruder from a passage's bottom 50\% of predicted topics by probability, and for TopicGPT-lite, we sample the intruder from topics not assigned to the passage by the model. Following \citet{bhatia-etal-2017-automatic}, we require intruders to be a top-ranked topic for at least one other document, to avoid ``junk'' or infrequent topics that would be trivial intruders. Annotators rate topics on an ordinal scale from 1 (\textit{Not Relevant}) to 3 (\textit{Very Relevant}) (full task instructions in Appendix~\ref{sec:appdx_passage}). We expect a good topic model's top predicted topic to be very or somewhat relevant to the passage, and intruders should be rated as not relevant. 

Since reading and understanding literature is time-intensive, this additional evaluation focuses on six approaches, spanning \texttt{Retell} and TopicGPT-lite, determined to be competitive in the previous evaluation. The approaches we select include those pertaining to one open and one closed LM: Llama 3.1 8B and GPT-4o mini. To ensure solid literary comprehension, we recruited two in-house annotators, who were not informed of our methodological approach but have prior experience annotating film and literature (Appendix~\ref{sec:appdx_passage}).  These annotators rated topics for 50 passages sampled from the dataset introduced in \S\ref{sec:eval_passage_sets}, with good agreement (weighted Cohen's $\kappa$ = 0.66).
%For each passage, we custom-write two comprehension checks (examples in Appendix~\ref{sec:appdx_passage}), and only consider ratings provided by annotators who passes both checks. 

% \paragraph{Qualitative analysis.} Finally, we conduct a sentence-by-sentence close reading of 30 passages and their summaries and descriptions outputted by Llama-3.1-8B and GPT-4o mini. Here, our goal is to identify retelling behaviors that may affect topic quality. Past literature on text summarization has pointed to faithfulness as a common weakness of LMs, where generated text may contain content inconsistent with or unsupported by the original text \cite{subbiah-etal-2024-storysumm, krishna2023longeval, kryscinski-etal-2020-evaluating}. So, we also mark unfaithful spans and code observations about them. 

% our two student annotators examine the summaries and descriptions produced by Llama-3.1-8B and GPT-4o mini for 30 passages. \todocomment{describe annotation task} Obtaining thorough faithfulness annotations is challenging; \todocomment{cite} showed that ``disagreements'' among experienced annotators can arise due to one annotator accidentally missing an error caught by another. Thus, we ask our student annotators to first annotate individually, and then discuss and produce a shared set of adjudicated annotations. 

\section{Evaluation Results} \label{sec:results}

\texttt{Retell}'s most prominently predicted topics for passages that share the same gold label demonstrate our approach's ability to surface cross-cutting themes in literary texts (Table~\ref{tab:example_topics}). \texttt{Retell}'s topics contain terms that tend to be more conceptual, e.g. \textit{financial}, than default LDA, e.g. \textit{money}. We also observe that TopicGPT-lite has a tendency to group a high proportion of passages under a few common, overly broad topics, e.g. Gemma's \textit{social interaction} (21.4\%) or Llama's \textit{life} (32.9\%). In Appendix~\ref{sec:automatic}, we compute methods' precision and recall of passage pairs that share the same gold topic, and find that \texttt{Retell} is generally more precise than baseline methods. Appendix~\ref{sec:automatic} also details how \texttt{Retell}'s higher precision generalizes across different gold label sources, and persists even when LMs are rerun. 

\texttt{Retell} produces topics that crowdworkers often judge to be semantically related to passage sets' gold labels (Table~\ref{tab:topic_relatedness}). By experimenting with different instructive verbs, we wished to see how abstractive verbs that encourage more ``telling'' (\textit{describe} and \textit{summarize}) may perform compared to a verb that may retain more ``showing'' (\textit{paraphrase}). As discussed in \S\ref{sec:intro}, we hypothesized that the former two should perform better than the latter. Averaged across all four models and choices of $k$ shown in Table~\ref{tab:topic_relatedness}, we indeed see that \texttt{Retell}-\textit{summarize} (\ding{51} $=$ 0.59) and \texttt{Retell}-\textit{describe} (\ding{51} $=$ 0.60) perform better than \texttt{Retell}-\textit{paraphrase} (\ding{51} $=$ 0.50). 

Still, there is some variation in how different models respond to different verbs. For instance, GPT-4o mini's summaries and descriptions outperform its paraphrases, while Gemma 2 2B's \texttt{Retell} variants do not share the same pattern, and its \texttt{Retell}-\textit{paraphrase} actually performs well (Table~\ref{tab:topic_relatedness}). Upon closer inspection of these LMs' retellings, we observe that GPT-4o's paraphrases rewrite the original passage in a more direct, low-level manner. Take for example a passage from from John Steinbeck's \textit{The Pearl}, which begins with the sentence \textit{Kino moved sluggishly, arms and legs stirred like those of a crushed bug, and a thick muttering came from his mouth}. GPT-4o's paraphrase of this passage starts with \textit{Kino moved slowly, his limbs resembling those of a squashed insect, as he muttered thickly to himself}. In contrast, Gemma 2 2B is less exact with instruction following, and its ``paraphrases'' sound more like summaries: \textit{This passage depicts the ripple effect of a single, extraordinary event--the discovery of a legendary pearl--on a small community}. Our observation suggests that though the \textit{paraphrase} vs. \textit{describe}/\textit{summarize} distinction may not be interpreted similarly by all LMs, downstream topic modeling performance may still relate to retellings' level of abstraction. 

Regardless of models' idiosyncrasies, our three variants of \texttt{Retell} consistently outperform the baselines default LDA and TopicGPT-lite (Table~\ref{tab:topic_relatedness}). Frequently predicted TopicGPT-lite labels, e.g. \textit{human nature}, are uninformative and imprecise, with more ratings as ``Somewhat Related'' than ``Very Related'' to gold labels. Across all methods, we find that the gold topics related to the social and behavioral concepts \texttt{hypocrisy}, \texttt{ambition}, and \texttt{innocence} are the most difficult to yield very related predictions, while the topics \texttt{war}, \texttt{religion}, and \texttt{education} are easiest. 

Human ratings of passage-level topic relevance provide additional support that \texttt{Retell} is a competitive alternative to TopicGPT-lite (Table~\ref{tbl:passage_eval}). Here, we compare a few of the more promising \texttt{Retell} approaches, as suggested by Table~\ref{tab:topic_relatedness}, against TopicGPT-lite. Our evaluation here includes one open model (Llama 3.1 8B) and one closed one (GPT-4o mini), and two values of $k$ for each: $k=50$ and the $k$ set by TopicGPT-lite. We note that it is not possible to perform apple-to-apple comparisons when comparing the top topics drawn from a probabilistic model (LDA-style outputs) with those outputted by a direct labeling approach (TopicGPT-lite). Still, on average, intruder topics have lower relevance scores than passages' top topics for both methods. Appendix~\ref{sec:appdx_passage} includes further discussion around the implications of differing topic formats for human interpretation. 

Across two forms of evaluation, we show that our proposed approach, \texttt{Retell}, is capable of identifying cross-cutting themes in literary passages. 

% \subsection{Qualitative Analysis} \label{sec:retell_analysis}

% A mixed-method analysis of retellings provides some insight into how models retell narrative passages. 
%We find that each LM's idiolect leads to some models systematically prefer some wordings over others (Appendix~\todocomment{TODO}). For example, \todocomment{example}. 
% Descriptions and summaries, which perform similarly (\S\ref{sec:results}), tend to be lexically similar. When it comes to stability, we find that Figure~\ref{fig:precision_recall}'s trends around \texttt{Retell}'s performance persist even when LMs are reprompted (Appendix~\todocomment{TODO}).

% We surface several observations around models' retellings. \todocomment{Currently Lucy is doing this ...  some discussion of types of faithfulness errors, LMs' use of prior knowledge about books likely learned during pretraining. If I make a taxonomy then the Appendix can include examples .................... .................... .................... .................... .................... .................... .................... .................... .................... .................... .................... .................... .................... .................... .................... .................... .................... .................... ....................} Overall, these observations point towards intriguing behaviors around LM-based literary summarization and interpretation for future work to investigate further.

\section{Case Study: Race in ELA Books}\label{sec:case_study}

Now that we've evaluated our approach across a variety of LMs and topics, we conduct a deep-dive demonstration and analysis of its ability to support distant reading within a specific use case. A common first-order task for scholars working with literature is identifying passages of interest for closer reading and analysis \cite{tangherlini2013trawling}. Bottom-up tagging of literary collections with topics can also support search and discovery \cite{Park_Brenza_2015}. Here, we investigate the extent to which topic modeling can surface passages of interest for researchers. Specifically, we examine whether predicted topics align with expert-guided, theory-driven human annotations. 

\begin{table}[t]
\centering
\resizebox{\columnwidth}{!}{
\begin{tabular}{lcccc}
\toprule
\textbf{Method} & \textbf{Topic 1} & \textbf{Topic 2} & \textbf{Topic 3} & \textbf{Intruder} \\
\midrule
\texttt{Retell}-\textit{summ.} \llama{}, $k=50$ & 2.36 & 2.30 & 2.12 & 1.67 \\
\texttt{Retell}-\textit{summ.} \llama{}, $k=206$ & 2.40 & 2.14 & 2.17 & 1.81 \\
TopicGPT-lite \llama{}, $k=206$ & 2.64 & 2.57 & 2.19 & 1.40 \\
\midrule
\texttt{Retell}-\textit{desc.} \gpt{}, $k=50$ & 2.81 & 2.51 & 2.23 & 1.63 \\
\texttt{Retell}-\textit{desc.} \gpt{}, $k=89$ & 2.60 & 2.53 & 2.40 & 1.77\\
TopicGPT-lite \gpt{}, $k=89$ & 2.59 & 2.48 & 2.51 & 1.52\\
\bottomrule
\end{tabular}
}
\caption{Passage-level human evaluation results in top topics' and intruders' ratings that differ significantly for each method ($U$-test, $p<0.05$). Ratings are on a scale of 1 (\textit{Not Relevant}) to 3 (\textit{Very Relevant}) and averaged across 50 passages. In the leftmost column, \llama{} = Llama 3.1 8B and \gpt{} = GPT-4o mini.}
\label{tbl:passage_eval}
\end{table} 

In particular, of ongoing interest in education and social psychology is the question of how race and racism is taught, or could be taught, to students \cite{perry2020initial, ladson_billings1995, lynn2006critical, Bedford02012023}. Within the context of English language arts (ELA), or courses that teach reading and writing skills, researchers may investigate how literature depicts race \citep[e.g.][]{adukia2023children}, how teachers contextualize these depictions \citep[e.g.][]{beach2015identity}, and how literature and pedagogical practice may combine to affect students' understanding of themselves and others \citep[e.g.][]{byrd2016does, cocco2021comparing}. To conduct these types of studies, it may be useful to identify literary passages within curricular collections that substantively engage with topics regarding racial/cultural identity. 

In this case study, we run topic modeling on 1,645 human-annotated passages drawn from books associated with ELA courses in the U.S. We supplement this set with 19.8k additional passages stratified sampled across these books, to yield robust topic estimates. We compare outputs produced by \texttt{Retell}-\textit{summarize} ($k=50$), TopicGPT-lite, and default LDA ($k=50$), with the former two run with an open LM, Llama 3.1 8B.\footnote{$k=100$ produced similar topics and conclusions.} By running three different methods juxtaposed against passage-level human annotations, we hope to illustrate how they may differ in practice. 

\begin{table*}[t]
\centering
\resizebox{\textwidth}{!}{
\begin{tabular}{p{3cm} | p{4cm}c | p{5.5cm}c | p{7.5cm}c}
\toprule  
\textbf{Set} & \textbf{TopicGPT-lite ($k$=223)} & \textbf{MRR} & \textbf{Default LDA ($k$=50)} & \textbf{Prob} &  \textbf{\texttt{Retell}-\textit{summarize} ($k$=50)} & \textbf{Prob} \\
\midrule
\multirow{3}{3cm}{Passages that only mention race (\texttt{mention})} 
& \textit{Identity} & 0.459 & 
\textit{black, people, white, new, american} & 0.041 & 
\textit{cultural, identity, heritage, family, complexities} & 0.062\\
& \textit{Family} & 0.357 & 
\textit{war, men, soldiers, two, army} & 0.033 & 
\textit{black, racial, white, community, individuals} & 0.034\\
& \textit{Work} & 0.356 & 
\textit{mother, father, family, children, son} & 0.032 & 
\textit{identity, self, complexities, explores, own} & 0.027\\
\midrule
\multirow{3}{3cm}{Passages that discuss race (\texttt{discuss})} 
& \textit{Identity} & 0.522 & 
\textit{black, people, white, new, american} & 0.085 & 
\textit{black, racial, white, community, individuals} & 0.106\\
& \textit{Relationships} & 0.352 & 
\textit{people, its, more, which, also} & 0.057 & 
\textit{cultural, identity, heritage, family, complexities} & 0.090\\
& \textit{Family} & 0.290 & 
\textit{think, want, maybe, really, how} & 0.045 & 
\textit{identity, self, complexities, explores, own} & 0.041\\
\midrule
\multirow{3}{3cm}{Other annotated passages (\texttt{neither})} 
& \textit{Identity} & 0.416 & 
\textit{think, want, maybe, really, how} & 0.029 & 
\textit{rather, than, human, not, can} & 0.031\\
& \textit{Work} & 0.361&  
\textit{says, has, can, does, say} & 0.028 & 
\textit{family, mother, father, dynamics, son} & 0.029\\
& \textit{Family} & 0.357 &  
\textit{people, its, more, which, also} & 0.028 & 
\textit{natural, world, landscape, nature, scene} & 0.027\\
\bottomrule
\end{tabular}
}
\caption{The top three most prominent topics in passages that only mention racial/cultural identity (\texttt{mention}), those that discuss it (\texttt{discuss}), and those that do not mention nor discuss it (\texttt{neither}), as defined in \S\ref{sec:case_study_data}. Topics are ordered by mean reciprocal rank (MRR) for TopicGPT-lite, and ordered by their average probability for default LDA and \texttt{Retell}.}
\label{tbl:race_discuss_topics}
\end{table*} 

\subsection{Data}\label{sec:case_study_data}

We draw passages from 396 books listed in \citet{lucy2025jca}'s study of racial/ethnic representation in high school ELA.\footnote{ELA booklists in the Cultural Analytics Dataverse: \url{https://doi.org/10.7910/DVN/WZQRRH}.} This data originates from two overlapping sources: 250 titles listed in AP Literature exams essay questions in 1999-2021 \cite{ap2024}, and 207 titles listed by 189 teachers who taught in schools with mainly Black and Hispanic student populations in 2016-2022 \cite{levine2021feeling}. This data spans two distinct views of ELA in the U.S. That is, AP Literature represents a standardized, prescriptive view of the literary canon \cite{abrams2023shortchanged}, while teacher-listed books demonstrate how curricula could potentially be adapted for diverse classrooms.

We focus on analyzing the topical composition of passages that \textit{explicitly} mention racial and/or cultural identity. Though race can be implicitly referenced, defining the scope of implicit cues is worthy of more in-depth study in future work, and so we focus on explicit mentions for our demonstrative study. Our view of what constitutes ``race'' draws from definitions established by educators, sociolinguists, and cultural analytics researchers \cite{bonilla2015, morning2011, milner2017, algee2020representing}. We include inherited categories as well as those shaped by shared cultural practices; that is, we consider references to ethno-religious and geographic origin groups, e.g. \textit{Jewish}, \textit{South Asian}, and racial ones, e.g. \textit{White}. 

We use a directed content analysis approach to construct a coding scheme for manually annotating passages that mention or discuss race \cite{hsieh2005three}, iteratively refining face-valid and theory-driven codes. Like in \S\ref{sec:eval}, passages are up to 250 words in length while respecting paragraph boundaries. In early stages of annotation, we observe that identifying passages of interest resembled searching for needles in a haystack. Thus, to raise our chances of coming across positive examples, we focus on annotating passages that include keywords in an extensive seed list (Appendix~\ref{appdx:case_study_ann}), including names of countries (e.g. \textit{Jamaica}), racial, ethnic, cultural, or nationality identifiers (e.g. \textit{Turkish}, \textit{Hmong}), or terminology related to racial issues (e.g. \textit{Whiteness}, \textit{undocumented}). 

Our annotator team consists of four undergraduates led by the second author, a social psychologist who specializes in research around racial identity.  Within four months, these annotators coded 1,645 passages. Annotators found 401 passages containing cursory mentions of racial identity (\texttt{mention}), and 198 involving deeper engagement with racial identity and issues (\texttt{discuss}), including descriptions of racial/cultural groups, instances of racially-based violence, negative racist language or dialogue, and mentions of racism or racial inequality. We consider passages that do not fall within \texttt{mention} nor \texttt{discuss} to be negative examples (\texttt{neither}). Appendix~\ref{appdx:case_study_ann} details our codebook, annotation process, and agreement levels, which were moderate (Cohen's $\kappa$ $\geq 0.5$) to substantial (Cohen's $\kappa$ $\geq 0.8$), depending on the code. Overall, combing through literature to identify passages of interest required much time and effort. 

\begin{figure*}[t]
    \includegraphics[width=0.38\textwidth]{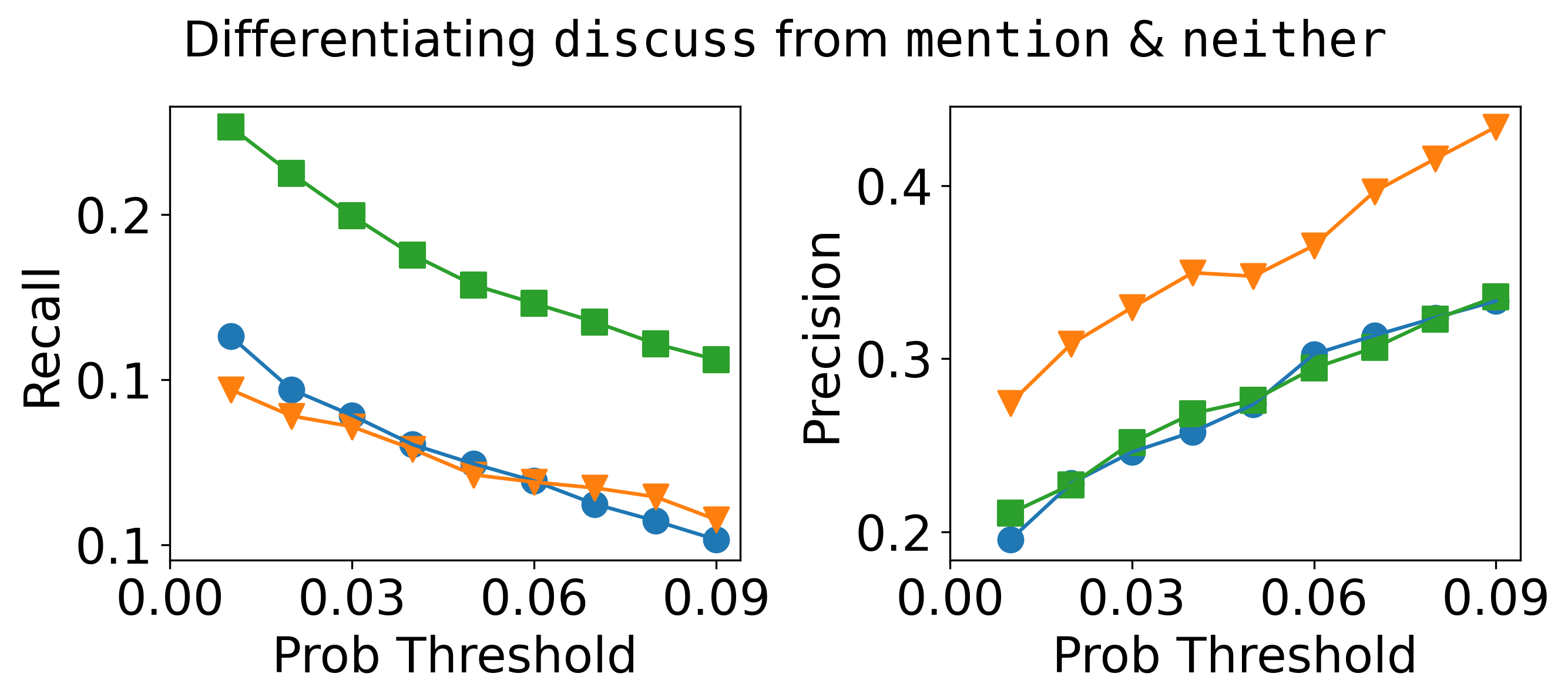}
    \includegraphics[width=0.38\textwidth]{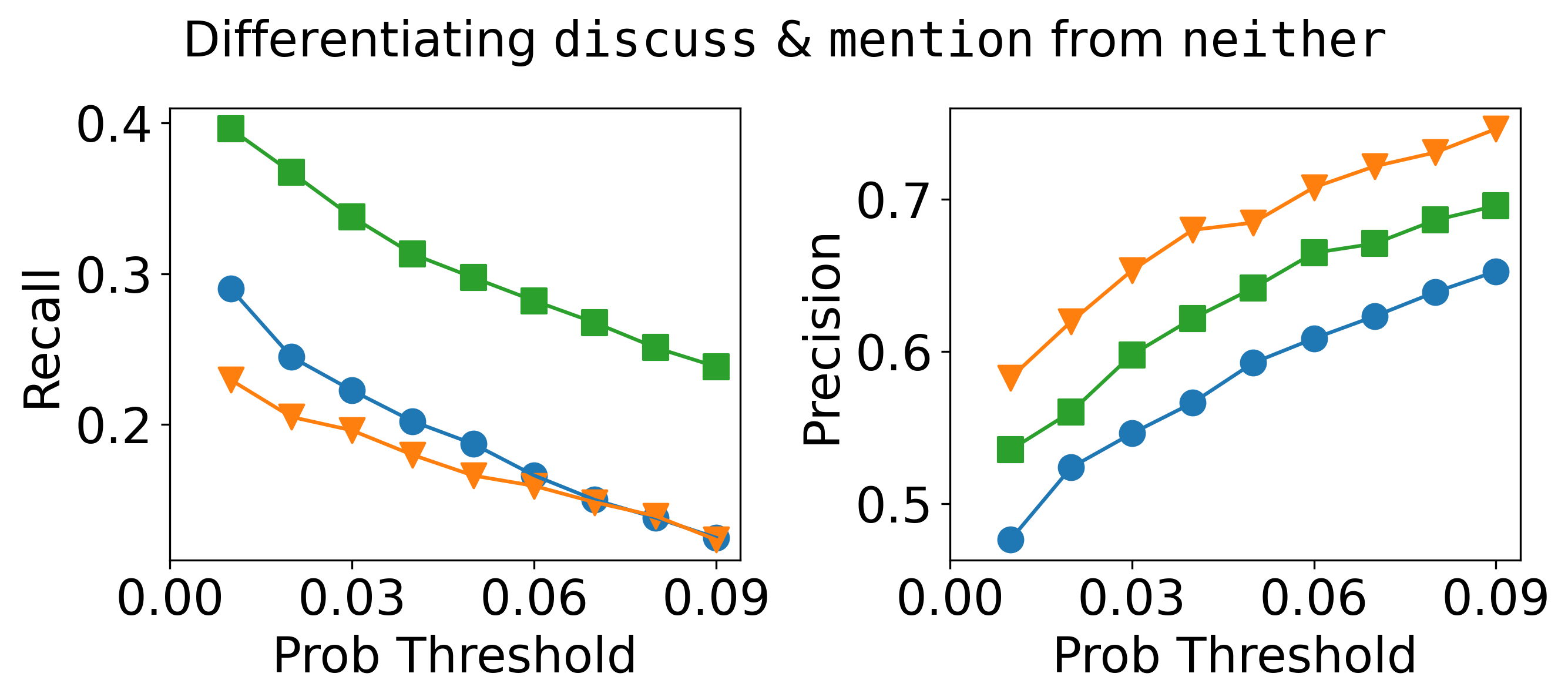}
    \raisebox{0.15\height}{
    \includegraphics[width=0.2\textwidth]{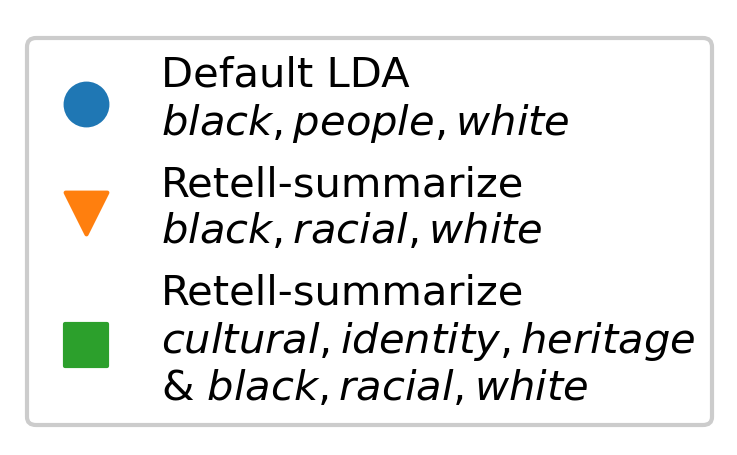}}
    % \begin{minipage}{0.65\columnwidth}
    % \includegraphics[width=\columnwidth]{figures/cs_2_vs_10.png}
    % \\
    % \includegraphics[width=\columnwidth]{figures/cs_21_vs_0.png}
    % \end{minipage}
    % \begin{minipage}{0.3\columnwidth}
    % \includegraphics[width=\columnwidth]{figures/cs_legend.png}
    % \end{minipage}
    \centering
    \caption{Precision and recall ($y$-axes) of using relevant \texttt{Retell} and LDA topics' probability thresholds ($x$-axes) to distinguish different types of human-labeled passages from each other (\texttt{discuss}, \texttt{mention}, \texttt{neither}). Examining \bluecircle{} against \orangetriangle{} compares only one topic per method, and \orangetriangle{} is more precise. Comparing \bluecircle{} with \greensquare{} considers all relevant topics each method surfaced, and \greensquare{} has higher recall without sacrificing precision. Left two plots: identifying passages that \texttt{discuss} race. Right two plots: identifying passages that either \texttt{discuss} or \texttt{mention} race.}
	\label{fig:precision_recall_cs}
\end{figure*}

\begin{table*}[t]
\centering
\resizebox{\textwidth}{!}{
\begin{tabular}{p{6cm}|p{7.5cm}p{5.6cm}p{2.5cm}}
\toprule  
\textbf{Passage excerpt} & \textbf{\texttt{Retell}-\textit{summarize}} & \textbf{Default LDA} & \textbf{TopicGPT-lite}  \\
\midrule
\midrule
\textit{... white families in one of the wealthiest cities in the country could hire colored domestics for as little as five dollars a week ...} --- \textit{The Warmth of Other Suns} by Isabel Wilkerson
& \makecell{\textit{financial, economic, family, reveals, life} \\ \hlcolor{\textit{black, racial, white, community, individuals}} \\ \textit{labor, work, animals, also, ability}} & \makecell{\textit{money, work, get, pay, dollars} \\ \textit{got, than, much, more, because} \\ \textit{people, its, more, which, also}} & \makecell{\textit{Work} \\ \textit{Poverty} \\ \textit{Security}}\\
\midrule
\textit{... these protesters are so much more vehement about Canadian-born Japanese ... Lots of people have been fired from their jobs. Business on Powell Street ... } --- \textit{Obasan} by Joy Kogawa
& \makecell{\hlcolor{\textit{black, racial, white, community, individuals}} \\ \textit{city, new, its, town, including} \\ \textit{financial, economic, family, reveals, life}} & \makecell{\textit{think, want, maybe, really, how} \\ \hlcolor{\textit{black, people, white, new, american}} \\ \textit{very, old, man, much, little}} & \makecell{\textit{Work} \\ \textit{Business} \\ \textit{Identity}} \\
\midrule
\midrule
\textit{Black boys are taught to replicate the white boy game ... I’ve played the game over and over again and have wounded black girls and women ...} --- \textit{No Ashes in the Fire} by Darnell L. Moore
& \makecell{\hlcolor{\textit{black, racial, white, community, individuals}} \\ \textit{societal, expectations, women, norms, social} \\ \textit{rather, than, human, not, can}} & \makecell{\textit{people, its, more, which, also} \\ \textit{play, music, game, playing, other} \\ \textit{which, men, these, great, man}} & \makecell{\textit{Relationships} \\ \textit{Identity} \\ \textit{Work}}\\
\midrule
\textit{There is something important about women in Guatemala, especially Indian women ... This feeling is born in women because of the responsibilities they have, which men do not have.} --- \textit{I, Rigoberta Menchú} by Rigoberta Menchú
& \makecell{\textit{societal, expectations, women, norms, social} \\ \hlcolor{\textit{cultural, identity, heritage, family, complexities}} \\ \textit{community, public, personal, including, system}} & \makecell{\textit{people, its, more, which, also} \\ \textit{woman, women, love, girl, wife} \\ \textit{which, life, its, love, heart}} & \makecell{\textit{Identity} \\ \textit{Family} \\ \textit{Work}} \\
\bottomrule
\end{tabular}
}
\caption{Illustrative excerpts from example passages and the top three topics predicted by each approach. These passages originate from the unlabeled pool of additional passages included in topic modeling (\S\ref{sec:case_study_data}). \texttt{Retell}-\textit{summarize} with Llama 3.1 8B relates the first two passages to socioeconomic status, and the last two to gendered social expectations. Our approach also relates all four passages to \hlcolor{racial and/or cultural identity}.}
\label{tbl:discuss_examples}
\end{table*} 

\subsection{Results \& Discussion}

\subsubsection{Topic Models' Behavior} \label{sec:cs_behavior}

We find that differences in passages' topical emphases relate to annotators' codes. That is, topics signaling racial/cultural identity are prominent in both \texttt{mention} and \texttt{discuss} passage sets (Table~\ref{tbl:race_discuss_topics}). For both default LDA and \texttt{Retell}-\textit{summarize}, the probability of topics that strongly suggest race, or \textit{black, people, white} and \textit{black, racial, white} for each approach respectively, are significantly higher in \texttt{discuss} than \texttt{mention}, matching our intentions in the design of these categories ($U$-test, $p<0.001$). In contrast, TopicGPT-lite's top topics do not signal race and/or culture, and are similar across all three passage sets (Table~\ref{tbl:race_discuss_topics}). Within the longer tail of TopicGPT-lite's $k=223$ labels, we do find three topics in its generated topic pool that are suggestive of race: \textit{Slavery}, \textit{Racism}, and \textit{Social Justice}. Though \textit{Social Justice} appears in predictions for 20.7\% of passages in \texttt{discuss}, it does not specify nor focus on racial/cultural identity. 

%Still, these labels do not specify nor focus on our target constructs. 
%We primarily investigate bottom-up, generalizable topic modeling methods, but this observation lends credence to more goal-directed content analysis approaches \citep[e.g.][]{wang-etal-2023-goal}. 

Next, we examine how various thresholds of topic probabilities compare to human annotations. That is, consider a simple classifier that predicts passages with one or more topics above some probability cutoff $c$ to be those that \texttt{discuss} and/or \texttt{mention} racial/cultural identity. Figure~\ref{fig:precision_recall_cs} shows that with \texttt{Retell}'s two relevant top topics, we're able to recover more passages of interest than with the one relevant top topic predicted by default LDA. That is, when \texttt{Retell}'s \textit{black, racial, white} is paired with a second prominent topic \textit{cultural, identity, heritage}, this approach can recall more passages than LDA's \textit{black, people, white} alone, without sacrificing precision. When considering only one topic for each method, \texttt{Retell}'s \textit{black, racial, white} can more precisely identify passages in \texttt{discuss} and \texttt{mention} than LDA's analogous topic can (Figure~\ref{fig:precision_recall_cs}). 

Though our case study focuses on racial and cultural identity, \texttt{Retell}'s general ability to surface a range of different topics (\S\ref{sec:results}) holds potential for the study of thematic co-occurrences. Table~\ref{tbl:discuss_examples} illustrates example passages where \texttt{Retell} topics related to racial and/or cultural identity intermix with other socially significant topics, including ones relevant for intersectional studies. With this table, one can also concretely see how each method's outputs differ at the passage-level, and the kinds of terms that compose the topics produced by each. 

\subsubsection{Error Analysis}

Though our approach shows promise compared to baselines, the performance in Figure~\ref{fig:precision_recall_cs} leaves room for improvement. To understand why Llama 3.1 8B's retellings may fall short, we qualitatively examine the 20 most severe false positives, or passages labeled as \texttt{neither} that have the highest probabilities of our two target topics. A few of these cases involve human annotators missing mentions of race, e.g. \textit{our rightful place ... in Black history}, suggesting a potential use of \texttt{Retell} as additional verification. In other cases, passages' discussion of racial/cultural identity is more implicit, falling outside the scope of our annotation codebook. For example, eight false positives discuss language use and differences, which the LM relates to culture or heritage. 

With some false positives, we observe that the LM's retelling may include additional book context that is not present in the original passage. For example, one retelling contrasts the passage's dialogue with a character's Iranian background, which is not mentioned in the passage but is accurately described by the LM. This output behavior arises from the presence of book summaries or the books themselves in pretraining data \cite{chang-etal-2023-speak, cooper2025extractingmemorized}. Like with human readers, an LM's interpretation of a passage may vary based on its familiarity with how the passage is situated within a larger narrative. Altogether, our observations around false positives suggest challenges around drawing a consistent divide between the implicit and explicit, as well as the passage-level and the book-level, when using LMs for literary text analysis. 

We also examine the 20 most severe false negatives, or passages labeled as \texttt{discuss} that have the lowest probabilities of our two target topics. Here, the LM's retellings may fail to recap content of interest, or insufficiently relate content to race/culture. For example, one retelling did not acknowledge the passage's description of a racist stereotype. An understudied challenge in text summarization research is determining what content is significant enough to recollect \citep[e.g][]{ladhak-etal-2020-exploring}, and these decisions may vary across different downstream tasks. Generally, we observe that classic summarization issues such as faithfulness and coherence \cite{zhang2024benchmarking, fabbri-etal-2021-summeval} do not severely degrade \texttt{Retell}. Instead, our observations point towards a need for more research characterizing LMs' \textit{content selection} behavior.

\section{Conclusion}

We propose a conceptually intuitive ``abstractive retelling'' approach for topic modeling of literature that pairs language models with classic LDA. Our proposed approach allows LDA to go beyond lexical-level sensory details in narrative text, and offers a competitive alternative to both LDA alone and directly generating topic labels, especially for resource-efficient, ``small'' LMs. We also examine our method's potential for search and exploration of literary collections in a case study on race in ELA books. Though \texttt{Retell} has room for improvement, it holds potential for supporting content analyses of literary collections at scale. Overall, we hope that the simplicity of our method encourages uptake in research around LMs' affordances and caveats in cultural analytics. 

\section*{Limitations} \label{sec:limitations}

Our framework proposes the idea that LMs can help us ``translate'' low-level narrative details (e.g. the clinking of glass at a table) into higher level concepts (e.g. a dinner scene). Though our method performs well against baselines, relying on LMs' ``retellings'' of passages can be reductive. Reading literature is a subjective, culturally constructed, and interpretive process \cite{fish1982there}, and LMs' retellings of passages typically only represent one interpretation. In the main text, we also touch on other shortcomings of LMs' retellings, including their failure to recap content of interest from the original text. 

Given our interdisciplinary audience's interests and constraints, we focus mainly on resource-efficient language models. This leaves open the question of how our approach may perform with larger LMs. We include some preliminary results comparing \texttt{Retell}-\textit{describe} and TopicGPT-lite with GPT-4o in Appendix~\ref{sec:appdx_gpt-4o}. There, we find that this stronger LM is proficient at directly outputting topic labels, so TopicGPT-lite achieves similar crowdsourced ratings of topic relatedness to ground truth topic labels as \texttt{Retell}. GPT-4o's TopicGPT-lite's distribution of passages across topics is also less skewed than that of smaller LMs (e.g. the most common top 3 topic, \textit{love}, appears in only 10.5\% of passages). These initial findings suggest that the task of outputting summaries/descriptions is easier than outputting topic labels directly, and so when a situation calls for smaller and/or weaker LMs, \texttt{Retell} provides a suitable alternative. We leave a more in-depth analysis for future work. Even if 2B to 8B sized LMs like the ones we tested improve, the million parameter models of the future may behave like the small billion parameter models of today. So, having options for topic modeling with different affordances (e.g. easily adjustable $k$) is still useful. 

We use gold labels scraped from online resources to evaluate our proposed method, allowing us to acquire many topics from both readers (Goodreads) and literature experts (LitCharts and SparkNotes). Still, we acknowledge that our use of a few select online resources risks content production biases, where content is affected by the context of a creator population \cite{olteanu2019social}. An issue of coverage also arises in any case of using naturally found data; online resources may not cover all possible tags/themes pertaining to a passage. So, we also include some human evaluation that does not use these gold labels (\S\ref{sec:passage_eval}, \S\ref{sec:case_study}).

% Finally, we note that the set of teachers in our case study are not representative of teachers who teach Black and Hispanic students in the U.S. as a whole \cite{lucy2025jca, levine2021feeling}. Mainly, our case study intends to illustrate how computational text analysis can support socially significant research involving literary text data. Our work only skims the surface of a broad line of research around culturally responsive curriculum design, which includes studies leveraging other research methods, e.g. interviews, to ground curriculum to educators' pedagogical decisions and practices and its impact on students. 

\section*{Ethical Considerations}

Our human evaluation in \S\ref{sec:eval} was deemed exempt from review by the institutional review board for human subjects research at our institution. 

Our case study in \S\ref{sec:case_study} codes passages mentioning race, and thus requires us to define and operationalize this abstract, socially constructed concept. Appendix~\ref{sec:appdx_cs_ann} describes how we scoped our annotations, including the definitions of race we consulted from prior work \cite{algee2020representing, bonilla2015, morning2011, milner2017}. For example, our annotation of race also includes ethnic and cultural categories. Though these constructs are interrelated, we acknowledge that race, ethnicity, and culture are distinct concepts \citep[e.g.][]{spencer2014race, worrell2014culture}. 

%The distinction between race and culture is not always straightforward to separate in language. For example, the term \textit{Chinese} could signify a cultural group, nationality, race, or multiple of the above. 

Our case study also proposes the use of LMs for identifying passages that discuss sensitive topics, e.g. racial/cultural identity. Extant social psychology and education research have shown that educational content and discussions about race, identity, and culture can be enriching and beneficial for all students, promoting empathy and critical thinking and reducing discrimination \cite{dee2017causal, brannon2021pride, apfelbaum2010blind}. Distant reading tools like ours should be used for \textit{supporting}, but not replacing, human researchers whose work intend to inform inclusive pedagogical practices. Our case study is mainly a proof-of-concept to further test our approach. Given the current sociopolitical climate in the U.S. around book banning and the teaching of race \cite{Alter_2024}, computational content analysis tools may be used to target books in harmful ways. LMs' descriptions and summaries of text, in their current form, should not be used for informing policy decisions, and the use of LMs in high stakes settings such as education requires careful collaboration with in-domain human experts.

\section*{Acknowledgements}

We thank Mackenzie Cramer and Anna Ho for providing human evaluation of passage-level topics (\S\ref{sec:passage_eval}), and Ariyanna Wesley, Kayla Collins, Kendal Murray, and Shevaun Yip for their efforts in coding racial/cultural identity for our case study (\S\ref{sec:case_study}). We also thank Julia Proshan for coordinating our coding process. 

\bibliography{custom}

\appendix

\section{Modeling Details}

\subsection{Retelling Parameters} \label{sec:appdx_params}

We run all open language models on 2 L40 GPUs, and download these models from Hugging Face. At the time of our experiments, \texttt{GPT-4o mini} in OpenAI's API pointed to \texttt{GPT-4o mini-2024-07-18}.

We run all LMs with \texttt{max\_new\_tokens} as 1024. To determine the ``default'' setups of each LM, we copied proposed settings in LMs' Hugging Face model cards or API documentation. For example, Phi-3.5-mini's Hugging Face model card recommends a temperature of 0. For LMs that allow a system prompt, which excludes Gemma 2, we used ``You are a helpful assistant; follow the instructions in the prompt.'' Our Github repository also includes code concretely showing how we run each model. 

\subsection{TopicGPT-lite}
\label{sec:apdx_topicgpt}

TopicGPT originally asked models to generate a topic hierarchy and assign topics within this hierarchy \cite{pham-etal-2024-topicgpt}. We simplify the task to only generate flat topics. During output parsing, we ignore any lines that do not follow our specified format of one topic per line, and during topic assignment, we disregard labels that were not introduced in the topic generation stage.

In TopicGPT, the original topic assignment prompt asks the model to support their assignment with evidence quoted from the passage. We remove this request from our prompts to reduce the difficulty of output formatting for smaller LMs. We also do not include refinement and correction stages in our adaptation of TopicGPT, which we constrain to require similar compute and model calls as \texttt{Retell}. That is, \citet{pham-etal-2024-topicgpt} propose reprompting the original LM if its outputs are not distinctive enough after topic generation, or if it hallucinates topics not in the provided list during topic assignment. Some models, such as Phi-3.5-mini, produces outputs that do not match the provided topic pool in 59.9\% of examples; reprompting to iteratively correct these issues would require numerous calls to the model, thus increasing the computational resource needs of this topic modeling approach. 

We run all TopicGPT-lite experiments using a fork of \citet{pham-etal-2024-topicgpt}'s code, from a version last updated by the authors on March 27, 2024. Figures~\ref{fig:topicgpt_generation}, \ref{fig:topicgpt_generation2}, \ref{fig:topicgpt_assignment} show our prompts for running TopicGPT-lite. These prompts were minimally edited from the original TopicGPT prompts, to encourage consistent instruction-following and output formatting from resource-efficient LMs. 

\section{Evaluation Data Collection}\label{sec:apdx_data}

\begin{table}[t]
\centering
\resizebox{\columnwidth}{!}{
\begin{tabular}{p{7cm}c}
\toprule  
\multicolumn{2}{c}{\textbf{Literary Passages \& Topic Labels}} \\
\midrule
\# of total passages for topic modeling & 23,332 \\
\hspace{10pt} \# of passages with gold topic labels & 11,666 \\
\midrule
Avg \# of white-spaced tokens per passage & 214.98 \\
Total \# of white-spaced tokens & 5.02M \\
\midrule
% \# of books across all passages & \todocomment{X} \\ 
% \hspace{10pt} \% of passages from Project Gutenberg & \todocomment{X} \\
% \hspace{10pt} \% of passages from contemporary booklists & \todocomment{X} \\
\# of books across gold-labeled passages & 732 \\ 
\hspace{10pt} \% of passages from Project Gutenberg & 25.6 \\
\hspace{10pt} \% of passages from contemporary booklists & 74.4 \\
\midrule
Avg \# of gold topics per labeled passage & 1.67 \\ 
\# of topic-passage pairs in total & 21,096 \\  
\hspace{10pt} \# labeled based on Goodreads tags & 9,072\\ 
\hspace{10pt} \# labeled based on Litcharts themes & 11,588\\
\hspace{10pt} \# labeled based on SparkNotes themes & 436 \\ 
\bottomrule
\end{tabular}
}
\caption{An overview of the labels and literary passages we use to evaluate topic modeling methods. The percentages from each book source do not total to 100 because they contain some overlapping titles.}
\label{tbl:data_stats}
\end{table} 

\begin{table}[t]
\centering
\resizebox{\columnwidth}{!}{
\begin{tabular}{l c c c}
\toprule  
\textbf{Topic} & \textbf{Goodreads} & \textbf{LitCharts} & \textbf{SparkNotes} \\
\midrule
\texttt{love} & 2269 & 789 & 40 \\
\texttt{literature \& stories} & 572 & 349 & - \\
\texttt{evil \& crime} & 429 & 1344 & 65 \\
\texttt{war} & 320 & 186 & 24 \\
\texttt{death} & 671 & 323 & 22 \\
\texttt{friendship} & 244 & 240 & - \\
\texttt{childhood, adulthood, \& age} & 353 & 530 & - \\
\texttt{religion} & 742 & 513 & 29 \\
\texttt{family} & 615 & 897 & 23 \\
\texttt{politics \& government} & 196 & 545 & - \\
\texttt{gender} & 739 & 1416 & 65 \\
\texttt{social class} & 234 & 782 & 52 \\
\texttt{time} & 146 & 145 & - \\
\texttt{race \& racism} & 767 & 1002 & 87 \\
\texttt{sex \& sexuality} & 83 & 253 & - \\
\texttt{forgiveness} & 57 & 148 & - \\
\texttt{education} & 244 & 114 & - \\
\texttt{language} & 151 & 383 & 29 \\
\texttt{innocence} & 31 & 144 & - \\
\texttt{justice} & 64 & 203 & - \\
\texttt{guilt} & 46 & 151 & - \\
\texttt{hypocrisy} & 40 & 99 & - \\
\texttt{morality} & 59 & 255 & - \\
\texttt{tradition} & - & 102 & - \\
\texttt{culture} & - & 471 & - \\
\texttt{loyalty} & - & 94 & - \\
\texttt{ambition} & - & 110 & - \\
\bottomrule
\end{tabular}
}
\caption{Number of passages within each gold topic label. Topic labels that do not have passages from that label source are indicated with a `-'.}
\label{tbl:topic_stats}
\end{table} 

\subsection{Books}

Our evaluation data encompasses both older, out-of-copyright titles and contemporary booklists. We obtain the former from Project Gutenberg, which provides already digitized full texts of books. For the latter, we include newer titles published in 1923-2021, digitizing them by scanning physical books. We use ABBYY Finereader for optical character recognition, and on a sample of excerpts from 20 books, we estimate character error rate to be on average 0.0049 (95\% CI [0.0002, 0.0096]).\footnote{Each excerpt is taken from the first page of books' main narratives, excluding paratext, and are at least 100 characters long, respecting paragraph boundaries.} We collect these contemporary booklists from a range of sources: Black-authored texts from the Black Book Interactive Project, non-U.S./U.K. global Anglophone fiction, Pulitzer Prize nominees, bestsellers reported by Publisher's Weekly and the New York Times, and \citet{lucy2025jca}'s high school ELA dataset. 
%Our contemporary dataset is similar to those found in prior work \cite{chang-etal-2023-speak, lucy-bamman-2021-gender}. 

% \begin{figure}[t]
%     \includegraphics[width=\columnwidth]{figures/precision_recall.png}
%     \includegraphics[width=\columnwidth]{figures/precision_recall_legend.png}
%     \centering
%     \caption{Precision and recall of \texttt{Retell} ($\bullet$) against baselines (\textbf{$\times$}) for Llama 3.1 8B (L), Phi-3.5-mini (P), Gemma 2 2B (G) and GPT-4o mini (O), for $k=50$ or various $k$ set by TopicGPT-lite. \texttt{Retell} can have higher precision than baseline approaches, and the high recall of TopicGPT-lite can lead to lower interpretive utility (\S\ref{sec:results}, Table~\ref{tab:topic_relatedness}). Performance patterns persist for $k=100, 200$ (Appendix~\ref{sec:appdx_eval}), and each point's 95\% CI is smaller than its marker size. }
% 	\label{fig:precision_recall}
% \end{figure}

\begin{figure}[t]
    \includegraphics[width=\columnwidth]{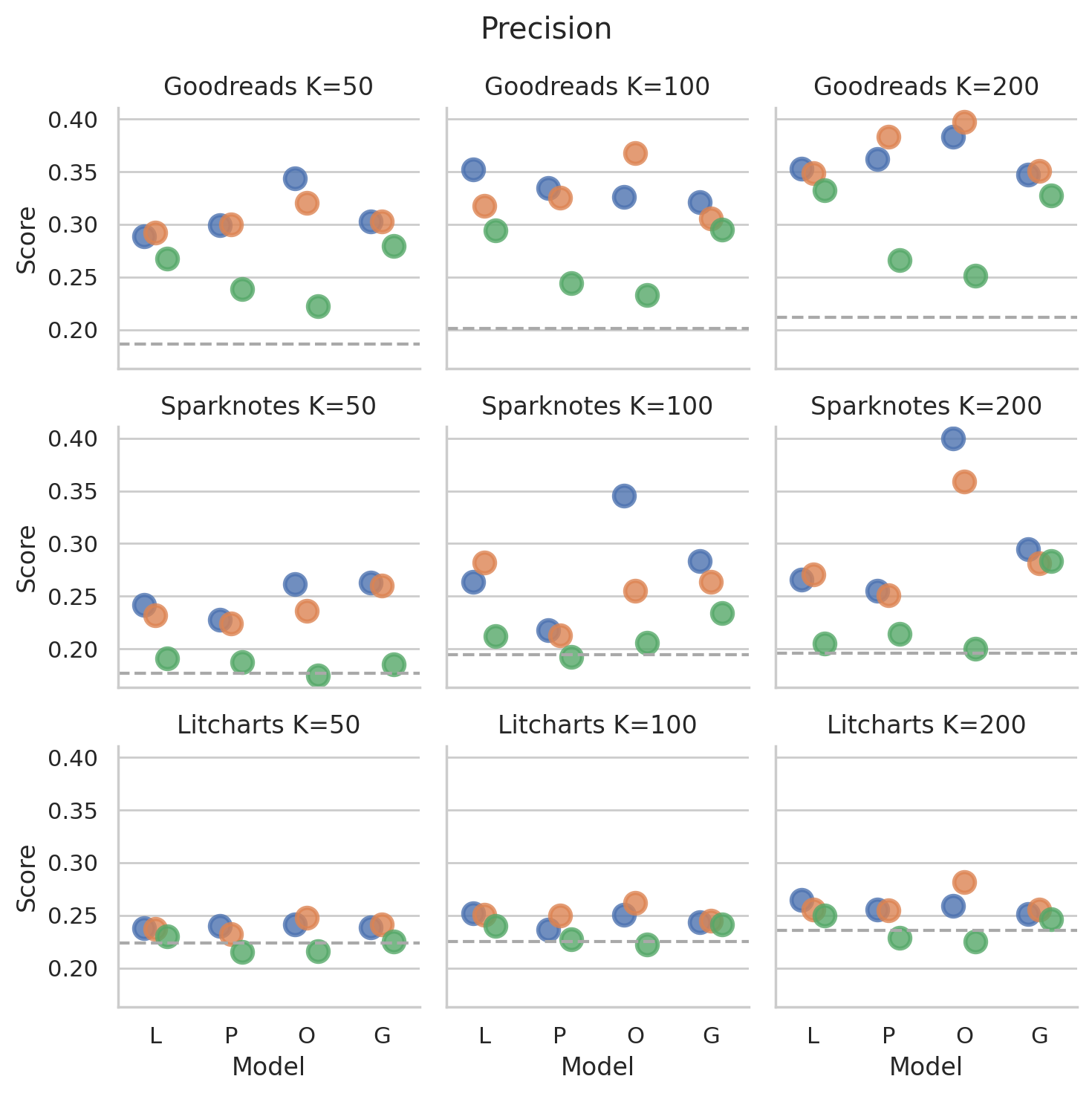}
    \includegraphics[width=\columnwidth]{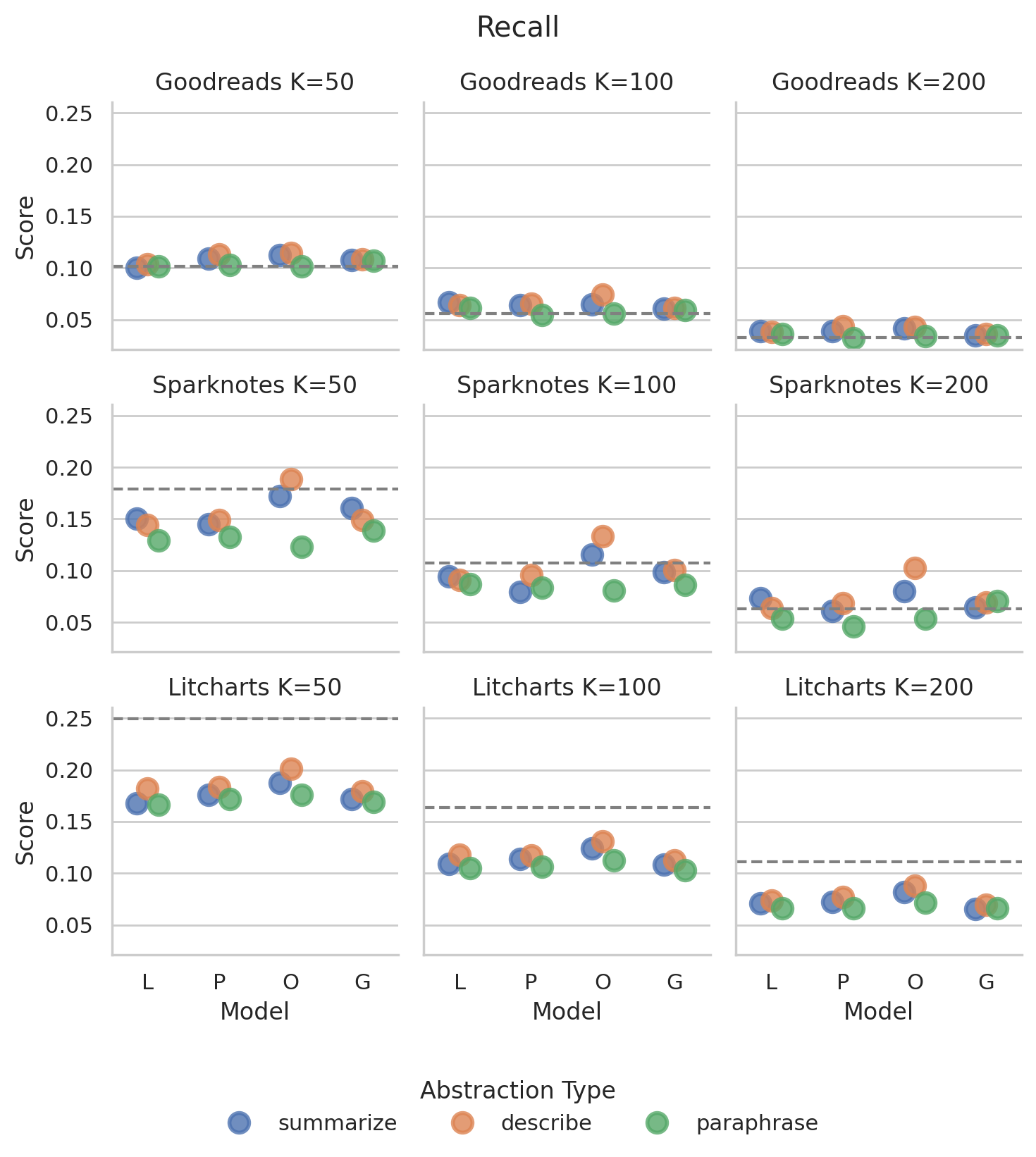}
    \centering
    \caption{Additional evaluation results for $k=50$, 100, and 200, disaggregated by the source of ground truth labels. Dashed gray lines indicate default LDA's performance, and the models on the $x$-axis include Llama 3.1 8B (L), Phi-3.5-mini (P), Gemma 2 2B (G) and GPT-4o mini (O). Since TopicGPT-lite is not designed for use with predetermined $k$, it is not included in these plots.}
	\label{fig:eval_by_source}
\end{figure}

\begin{figure}[t]
    \includegraphics[width=\columnwidth]{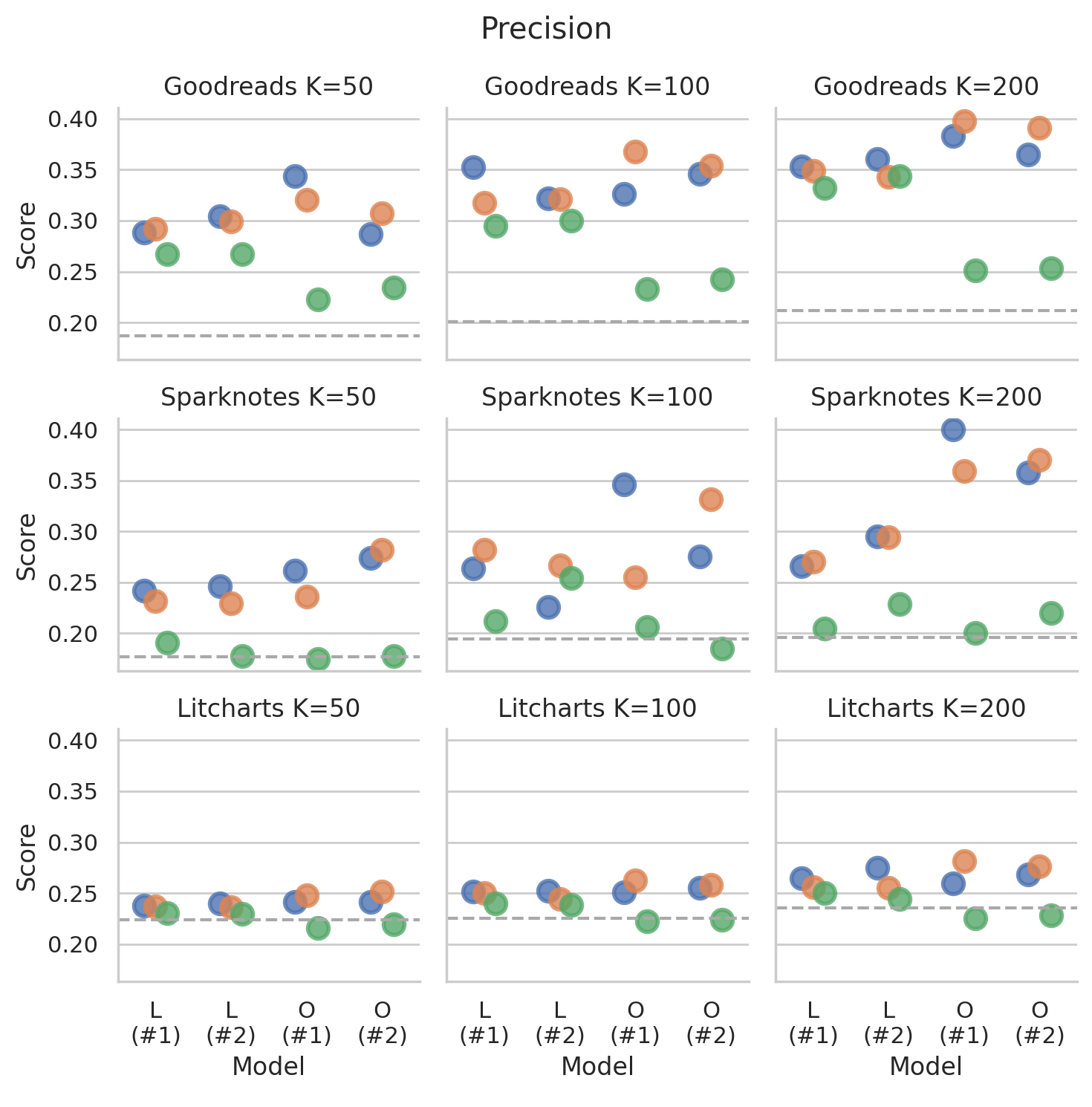}
    \includegraphics[width=\columnwidth]{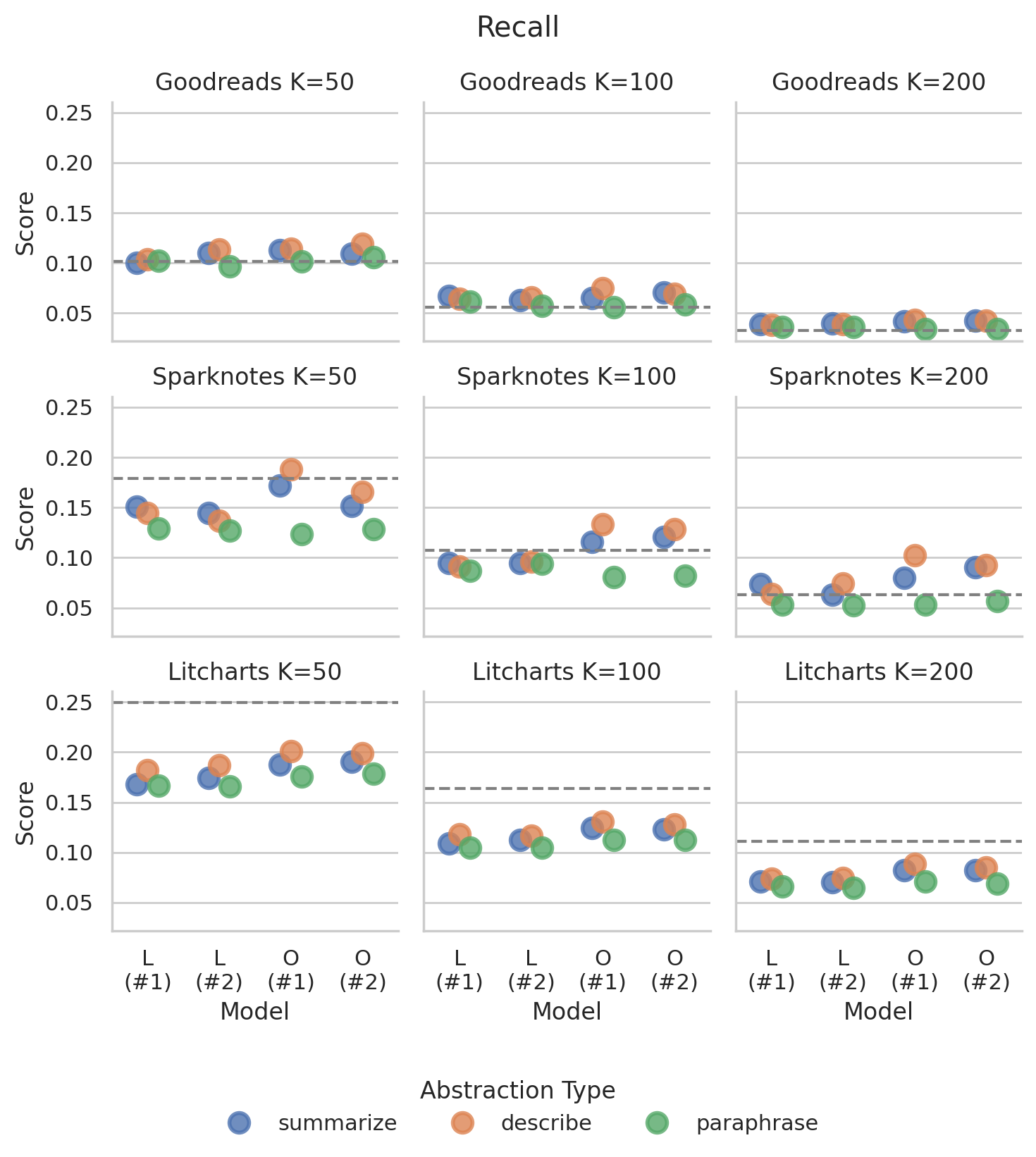}
    \centering
    \caption{Results for two different runs of Llama 3.1 8B (L) and GPT-4o mini (O). Dashed gray lines indicate default LDA's performance.}
	\label{fig:eval_by_run}
\end{figure}

\subsection{Themes, Tags, \& Topics}

% Matching titles in our book collection to each site yielded 1172 books' Goodreads pages, 508 books' LitCharts pages, and 417 books' SparkNotes pages. 

We match books in our collection to their pages on Goodreads, LitCharts, and SparkNotes, and scrape for tags/themes that accompany quoted material. We matched quotes in online resources to passages in books using fuzzy matching, where the normalized indel similarity between the two should exceed 90, on a scale from 0 to 100. Indel similarity considers the minimum number of insertions or deletions between two sequences. Our matching process achieved a match rate of 68.0. Common failure cases include misquotes, such as the quoted content missing a significant portion of the original book excerpt, or quotes that contain ``...'' signaling a portion of the excerpt was deliberately excluded. Qualitatively, we observed that the preceding passage leading up to a quote often shared similar themes as the quote, with the quote acting like a ``punch line'' to the passage as whole. Table~\ref{tbl:data_stats} includes a breakdown of our evaluation dataset, and Table~\ref{tbl:topic_stats} includes our full list of topics per label source and the number of passages with each label.

Our evaluation dataset is a multi-label one, as some quotes have multiple tags/themes, and some tags/themes are recoded into multiple topics. That is, passages originally labeled with \textit{interracial relationships} belong to both \texttt{love} and \texttt{race}. We create our set of gold topic labels by manually reviewing all Goodreads tags that occur at least 50 times among our books and all Litcharts and Sparknotes themes. To reduce the cognitive load of sifting through a messy pile of labels, we sorted tags/themes by words in them during the recoding process (e.g. the themes \textit{women and gender} and \textit{men and women} would be listed together since they share the word \textit{women}). We exclude tags/themes that did not fall within our final set of 27 gold topic labels, and also exclude Goodreads tags that did not relate to topic, e.g. author-related tags (\textit{john-green}) or tags suggesting meta-level reader commentary (\textit{inspirational}). 

\section{Evaluation} \label{sec:appdx_eval}

\subsection{Automatic Metrics} \label{sec:automatic}

One way to quantify topic modeling behavior across a range of LMs is to compute the precision and recall of passage pairs linked by each gold label source, e.g. Sparknotes \texttt{family} or Goodreads \texttt{religion}. We calculate these metrics in a manner that accounts for our multi-label data. 

\paragraph{Precision.} Given a passage $i$ labeled with an unordered set of $n$ gold topics $y_{i1}, y_{i2}, ..., y_{in} \in Y_i$, a topic modeling method predicts the passage's $p$ topics in order of most to least prominent: $\hat{y}_{i1}$, $\hat{y}_{i2}$, ... $\hat{y}_{ip}$. For LDA-based methods, $p = k$ since each passage has probabilities calculated across all topics, while for TopicGPT-lite, a model may output variable $p$ for each passage, where $p < k$. To calculate precision, we calculate the fraction of predicted positives, which are passage pairs $i, j$ with $\hat{y}_{i1} = \hat{y}_{j1}$, that share at least one gold topic label, or $Y_i \cap Y_j \neq \varnothing$. 

\paragraph{Recall.} We calculate recall based on a passage $i$'s $m$ unordered gold tag/theme labels $z_{i1}, ..., z_{im} \in Z_i$. These tags/themes are unique to each source, e.g. \textit{unrequited-love} in Goodreads, and they are finer-grained than gold topic labels, e.g. \texttt{love}. Using tags/themes allows us to avoid over-penalizing the failure to recall broad topics with higher $k$. Considering our multi-label setting, we compute recall with passages' top three $\hat{y}$. That is, we calculate the fraction of passage pairs $i, j$ with $Z_i \cap Z_j \neq \varnothing$ that have $\{\hat{y}_{i1}, \hat{y}_{i2}, \hat{y}_{i3}\} \cap \{\hat{y}_{j1}, \hat{y}_{j2}, \hat{y}_{j3}\} \neq \varnothing$. 

% In addition to these two pairwise metrics, we include results for two cluster-level metrics, purity and inverse purity, in Appendix~\ref{sec:appdx_purity} \todocomment{TODO}. Purity and inverse purity are analogous to precision and recall, respectively, and yield similar method performance patterns as our pairwise metrics. 

% \subsection{Cluster Purity and Inverse Purity} \label{sec:appdx_purity}

% \paragraph{Purity.} Purity is a common metric for evaluating document clustering \cite{schutze2008introduction}. 

\begin{table*}[h]
\centering
% \footnotesize
\def\arraystretch{0.8}
\setlength{\tabcolsep}{0.3em}
\resizebox{\textwidth}{!}{
\begin{tabular}{c | ccc cccccc ccccccccc ccccccccc}
\toprule
 & \multicolumn{6}{c}{Default LDA} & \multicolumn{3}{c}{TopicGPT-lite} & \multicolumn{9}{c}{\texttt{Retell}-\textit{describe}} & \multicolumn{9}{c}{\texttt{Retell}-\textit{summarize}} \\ 
\cmidrule(lr){1-2} \cmidrule(lr){2-7} \cmidrule(lr){8-10} \cmidrule(lr){11-19} \cmidrule(lr){20-28}

 & \multicolumn{6}{c}{\textit{no language model}} & \multicolumn{2}{c}{O} & L & \multicolumn{5}{c}{O} & \multicolumn{4}{c}{L} & \multicolumn{5}{c}{O} & \multicolumn{4}{c}{L} \\ 
\cmidrule(lr){1-2} \cmidrule(lr){2-7} \cmidrule(lr){8-9} \cmidrule(lr){10-10} \cmidrule(lr){11-15}
\cmidrule(lr){16-19} \cmidrule(lr){20-24} \cmidrule(lr){25-28}

\textbf{Rating} & 14 & 89 & 206 & 50 & 100 & 200 
& 14 & 89 & 206
& 14 & 89 & 50 & 100 & 200 & 206 & 50 & 100 & 200 
& 14 & 89 & 50 & 100 & 200 & 206 & 50 & 100 & 200  \\
\midrule
\ding{51} & \gradient{0.07} & \gradient{0.22} & \gradient{0.20} & \gradient{0.03} & \gradient{0.25} & \gradient{0.20}
& \gradient{0.30} & \gradient{0.28} & \gradient{0.48}
& \gradient{0.38} & \gradient{0.67} & \gradient{0.68} & \gradient{0.68} & \gradient{0.67}
& \gradient{0.55} & \gradient{0.53} & \gradient{0.48} & \gradient{0.60}
& \gradient{0.20} & \gradient{0.65} & \gradient{0.60} & \gradient{0.60} & \gradient{0.67}
& \gradient{0.67} & \gradient{0.58} & \gradient{0.60} & \gradient{0.63} \\
\textbf{?} & \gradient{0.18} &\gradient{0.20} & \gradient{0.12} & \gradient{0.12} & \gradient{0.13} & \gradient{0.20}
& \gradient{0.28} & \gradient{0.30} & \gradient{0.42}
& \gradient{0.37} & \gradient{0.22} & \gradient{0.20} & \gradient{0.23} & \gradient{0.27}
& \gradient{0.35} & \gradient{0.40} & \gradient{0.37} & \gradient{0.22}
& \gradient{0.60} & \gradient{0.22} & \gradient{0.30} & \gradient{0.30} & \gradient{0.27}
& \gradient{0.20} & \gradient{0.33} & \gradient{0.25} & \gradient{0.25} \\
\ding{55} & \gradient{0.75} & \gradient{0.58} & \gradient{0.68} & \gradient{0.85} & \gradient{0.62} & \gradient{0.60} 
& \gradient{0.42} & \gradient{0.42} & \gradient{0.10}
& \gradient{0.25} & \gradient{0.12} & \gradient{0.12} & \gradient{0.08} & \gradient{0.07} 
& \gradient{0.10} & \gradient{0.07} & \gradient{0.15} & \gradient{0.18}
& \gradient{0.20} & \gradient{0.13} & \gradient{0.10} & \gradient{0.10} & \gradient{0.07} 
& \gradient{0.13} & \gradient{0.08} & \gradient{0.15} & \gradient{0.12} \\
\bottomrule
\end{tabular}
}
\caption{An version of Table~\ref{tab:topic_relatedness} containing crowdsourcing results for more values of $k$ and one open LM (Llama 3.1 8B) and one closed one (GPT-4o mini).}
\label{tab:topic_relatedness_extended}
\end{table*}

\begin{table}[t]
\centering
\resizebox{\columnwidth}{!}{
\begin{tabular}{l | c l c c}
\toprule  
\textbf{Model} & \textbf{\textit{k}} & \textbf{Method} & \textbf{Precision} & \textbf{Recall} \\
\midrule
 & 50 & \texttt{Retell}-\textit{summarize} & 0.128 & 0.103 \\
 & 50 & \texttt{Retell}-\textit{describe} & \textbf{0.128} & 0.107 \\
 & 50 & \texttt{Retell}-\textit{paraphrase} & 0.121 & 0.105 \\ 
 & 50 & default LDA (no LM) & 0.100 & \textbf{0.108} \\
 \cmidrule{2-5}
Llama 3.1 8B & 206 & \texttt{Retell}-\textit{summarize} & \textbf{0.155} & 0.040 \\
 & 206 & \texttt{Retell}-\textit{describe} & 0.149 & 0.039 \\
 & 206 & \texttt{Retell}-\textit{paraphrase} & 0.135 & 0.036 \\
 & 206 & default LDA (no LM) & 0.107 & 0.036 \\
 & 206 & TopicGPT-lite & 0.127 & \textbf{0.211} \\
\midrule
 & 50 & \texttt{Retell}-\textit{summarize} & \textbf{0.131} & 0.112 \\
 & 50 & \texttt{Retell}-\textit{describe} & 0.131 & 0.\textbf{116} \\
 & 50 & \texttt{Retell}-\textit{paraphrase} & 0.110 & 0.106 \\
 & 50 & default LDA (no LM) & 0.100 & 0.108 \\
\cmidrule{2-5}
Phi-3.5-mini & 538 & \texttt{Retell}-\textit{summarize} & \textbf{0.170} & 0.020 \\
 & 538 & \texttt{Retell}-\textit{describe} & 0.159 & 0.021 \\
 & 538 & \texttt{Retell}-\textit{paraphrase} & 0.128 & 0.017 \\
 & 538 & default LDA (no LM) & 0.114 & 0.018 \\
 & 538 & TopicGPT-lite & 0.139 & \textbf{0.048} \\
\midrule
 & 50 & \texttt{Retell}-\textit{summarize} & 0.131 & 0.111 \\
 & 50 & \texttt{Retell}-\textit{describe} & \textbf{0.133} & \textbf{0.111} \\
 & 50 & \texttt{Retell}-\textit{paraphrase} & 0.123 & 0.110 \\
 & 50 & default LDA (no LM) & 0.100 & 0.108 \\
  \cmidrule{2-5}
Gemma 2 2B & 157 & \texttt{Retell}-\textit{summarize} & \textbf{0.147} & 0.045 \\
 & 157 & \texttt{Retell}-\textit{describe} & 0.145 & 0.046 \\
 & 157 & \texttt{Retell}-\textit{paraphrase} & 0.138 & 0.044 \\
 & 157 & default LDA (no LM) & 0.104 & 0.042 \\
 & 157 & TopicGPT-lite & 0.139 & \textbf{0.201} \\
\midrule
 & 50 & \texttt{Retell}-\textit{summarize} & \textbf{0.141} & 0.116 \\
 & 50 & \texttt{Retell}-\textit{describe} & 0.137 & \textbf{0.118} \\
 & 50 & \texttt{Retell}-\textit{paraphrase} & 0.106 & 0.105 \\
 & 50 & default LDA (no LM) & 0.100 & 0.108 \\
  \cmidrule{2-5}
 & 14 & \texttt{Retell}-\textit{summarize} & 0.110 & 0.244 \\
 & 14 & \texttt{Retell}-\textit{describe} & 0.118 & 0.243 \\
GPT-4o mini & 14 & \texttt{Retell}-\textit{paraphrase} & 0.100 & 0.226 \\
 & 14 & default LDA (no LM) & 0.092 & 0.222 \\
 & 14 & TopicGPT-lite & \textbf{0.127} & \textbf{0.248} \\
  \cmidrule{2-5}
 & 89 & \texttt{Retell}-\textit{summarize} & 0.141 & 0.079 \\
 & 89 & \texttt{Retell}-\textit{describe} & \textbf{0.154} & 0.082 \\
 & 89 & \texttt{Retell}-\textit{paraphrase} & 0.112 & 0.068 \\
 & 89 & default LDA (no LM) & 0.105 & 0.069 \\
 & 89 & TopicGPT-lite & 0.148 & \textbf{0.186} \\
\bottomrule
\end{tabular}
}
\caption{Precision and recall of \texttt{Retell} against baselines for $k=50$ or various $k$ set by TopicGPT-lite. \texttt{Retell} can have higher precision than baseline approaches, and the high recall of TopicGPT-lite can lead to lower interpretive utility (\S\ref{sec:results}, Table~\ref{tab:topic_relatedness}). The largest 95\% CI over 1k bootstrapped passage pairs for any value in this table is $\pm 8.0 \times 10^{-5}$, and since our results are rounded to the thousandth, we do not include CI ranges in this table. Highest values for each LM and choice of $k$ are bolded.}
\label{tbl:precision_recall}
\end{table} 

\paragraph{Results.} Linking passage pairs with shared themes is a difficult task; abstractive retelling can be more precise than baseline methods, but has similar or lower recall (Table~\ref{tbl:precision_recall}). As discussed in the main text (\S\ref{sec:eval}), high recall for methods like TopicGPT-lite arise from this approach labeling many passages with the same label, e.g. \textit{human nature}. Though these common labels may be generally relevant to passages, they are not always informative or distinctive. Our automatic metric results do support our \textit{tell} vs. \textit{show} motivation; \texttt{Retell}-\textit{summarize} and \texttt{Retell}-\textit{describe} tend to perform better than \texttt{Retell}-\textit{paraphrase} across all four LMs.

Figure~\ref{fig:eval_by_source} breaks down precision and recall results for \texttt{Retell} by label source, providing additional confirmation of the aggregated trends of Table~\ref{tbl:precision_recall}. That is, \texttt{Retell}-\textit{describe} and \texttt{Retell}-\textit{summarize} perform better than \texttt{Retell}-\textit{paraphrase}. Generally, descriptions and summaries are more similar to each other than they are to paraphrases. The Jensen-Shannon divergence between the word distributions of each LM's descriptions and summaries ($M=0.012$, $SD = 0.004$) is lower than that of paraphrases vs. summaries ($M=0.029$, $SD=0.021$) and paraphrases vs. descriptions ($M=0.048$, $SD=0.034$). 

Past work on topic modeling has highlighted stability issues in some methods, where small changes in parameters or rerunning the same model may yield consequentially different results \cite{hoyle-etal-2022-neural}. Two of our LMs, GPT-4o mini and Llama 3.1 8B, use default parameters where they are run with non-zero temperatures and non-greedy decoding, and so they produce different retellings when reprompted. Figure~\ref{fig:eval_by_run} shows that there is some variation between two different runs of \texttt{Retell}. Still, \texttt{Retell}-\textit{summarize} and \texttt{Retell}-\textit{describe} still typically have better performance than \texttt{Retell}-\textit{paraphrase}, and for precision, the former two consistently outperform default LDA.  

% We find that between two runs of these LMs, \todocomment{X}\% of passage pairs $i, j$ with $\hat{y}_{i1} = \hat{y}_{j1}$ in the first run are again paired together in the second one \todocomment{maybe make this into a table for each non-determistic LM and retelling approach}. 

\subsection{Topic Relatedness Crowdsourcing} \label{sec:appdx_topic_rel}

In \S\ref{sec:eval_passage_sets}, we compare passage sets' most prominently predicted labels with ground truth ones. We recruited Prolific workers based in the U.S. whose first language is English, with an approval rate between 95 and 100 and at least 10 prior submissions. Across all method and parameter combinations shown in Table~\ref{tab:topic_relatedness}, \ref{tab:topic_relatedness_extended}, and \ref{tbl:gpt_4o}, crowdworkers annotated 2940 pairs of predicted and ground truth labels, as each approach involved annotating 60 passage sets. We calculate inter-annotator agreement on 304 doubly annotated examples. Workers annotated examples in batches of up to 12 task examples at a time, where one example may be doubly annotated with another batch, and one example was a quality check example. We recollected ratings in cases where workers failed this check. 

These quality check examples were sampled from the following 20 possibilities, with the correct rating in parentheses and an ampersand delineating the faux gold and predicted topic labels, respectively: sports \& sports (Very Related), animals \& literature (Not Related), sports \& pitcher, sports, inning, player, bat (Very Related), animals \& table, couch, lamp, rug, coffee (Not Related), nature \& the natural world (Very Related), fruits and vegetable \& spaceship (Not Related), farm \& cow, chicken, farmer, farm, barn (Very Related), vehicle, \& asleep, nap, dreamy, pillow, sleep (Not Related), hatred \& hate (Very Related), kindness \& lawn, grass, weeds, bee, garden (Not Related), hygiene \& toothbrush, dental, shower, wellness, wash (Very Related), technology \& fruit (Not Related), sports \& athletics (Very Related), animals \& crisis (Not Related), vehicle \& driving, motor, car, vehicle, transport (Very Related), kindness \& bakery (Not Related), technology \& technology (Very Related), hygiene \& be, the, can, if, will (Not Related), farm \& farming and rural life (Very Related), hatred \& cat, dog, pet, fish, hamster (Not Related). 

Our full task instructions are the following: \\

\noindent \textit{Rate how much a word, phrase, or list of words is related to a given theme. There are three options: ``Very related", ``Somewhat related", and ``Not related".}  \\

\noindent \textit{Note: ``very related" includes cases where the provided word/s contains the theme.}  \\

\noindent \textit{Examples:}  \\

\noindent \textit{Theme: nature}

\noindent \textit{Word/s or Phrase: environment} 

\noindent \textit{Answer: Very related} \\

\noindent \textit{Theme: nature}

\noindent \textit{Word/s or Phrase: biotechnology}

\noindent \textit{Answer: Somewhat related}  \\

\noindent \textit{Theme: nature}

\noindent \textit{Word/s or Phrase: table}

\noindent \textit{Answer: Not related}  \\

\noindent \textit{Theme: nature}

\noindent \textit{Word/s or Phrase: tree, lake, mountain, animals, nature}

\noindent \textit{Answer: Very related}  \\

\noindent \textit{Theme: nature}

\noindent \textit{Word/s or Phrase: biology, cells, organism, microscope, experiment}

\noindent \textit{Answer: Somewhat related}  \\

\noindent \textit{Theme: nature}

\noindent \textit{Word/s or Phrase: computer, keyboard, mouse, wifi, hardware}

\noindent \textit{Answer: Not related}   \\

Then, for each predicted and gold topic label pair, we asked: \\

\noindent \textit{How related are the given word/s or phrase to the theme?}

\noindent \textit{Theme:} \texttt{gold label}

\noindent \textit{Word/s or Phrase:} \texttt{predicted label}

\noindent \subsection{Passage-level Topic Annotation} \label{sec:appdx_passage}

Our two annotators for this passage-level evaluation were finishing bachelor's degrees in rhetoric, English, data science, and cognitive science when conducting their annotations. Each had approximately two years of experience working with cultural analytics datasets, including literature and film. They annotated 50 passage and topic sets in text files, with one file per passage. Within each passage's annotation file, the passage itself was repeated above each of six sets of four topics to reduce scrolling. Annotators were not told how the topics were generated or that only one topic was an ``intruder''. 

We included the following instructions: 

\noindent \textit{In this task, you will read a literary passage. Try your best to understand the themes and concepts in the passage thoroughly.}

\noindent \textit{Then, you will rate the relevance of topics, or themes, to a passage. For each rating, you have three options: 3 - Very Relevant, 2 - Somewhat Relevant, and 1 - Not Relevant. Type your judgement within 1 to 3 next to each topic (see example below). }

\noindent \textit{A relevant topic should represent prominent themes or concepts in the passage. Each topic is either represented by a list of words (e.g. ``school, students, education, high, student"), or by a single word or phrase (e.g. ``Relationships"). You will be presented with four topics at a time, and there are six sets of them in total.}

\noindent \textit{You may ignore errors caused by the book's digitization process, which may create misspellings, extraneous letters or numbers, or unusual line/paragraph breaks.}

At the end of these instructions, we include one example passage from \textit{Always Running: La Vida Loca} by Luis J. Rodriguez, along with four labeled \texttt{Retell} topics, to show annotators how to format their responses. 

Annotators doubly annotated ten passages (ratings for 240 topics in total) when calculating inter-annotator agreement (IAA). The two annotators and the first author discussed the topic options for one passage, to calibrate our expectations of what ratings of ``Very Relevant'', ``Somewhat Relevant'', and ``Not Relevant'' may entail. During this discussion, we acknowledged that this task involves subjective or inferential interpretation of literary text. That is, the two annotators may have \textit{valid} disagreements when relating topics to passages, and thus should follow their own intuition while annotating, rather than rate what they expect the other annotator to rate solely for increasing IAA. 

When annotating the first 10 passages, annotators also asked the first author clarifying questions around the instructions. For example, one annotator initially stated that a military topic was \textit{not} relevant to a passage from \textit{The Three Musketeers} by Alexandre Dumas, because the musketeer characters are \textit{private} soldiers rather than formally part of the military. Initial discussions over cases of uncertainty like these were helpful for calibrating the scale of what is or is not relevant, and that topics may cover broad concepts (e.g. ``military'' may not differentiate different types of soldiers). 

Our annotators also made other observations during this task. For example, some topics (which in some cases were intruders) included in \texttt{Retell} topic sets were uninformative, e.g. ``\textit{can, who, than, how, rather}'' do not convey much actual content. Default LDA, which we did not annotate at the passage-level, is especially liable to producing stopword-esque topics like these (e.g. Table~\ref{tab:example_topics}), and though \texttt{Retell} is less prone to this issue, it still occurs. In addition, some \texttt{Retell} topics may be difficult to interpret if one is unfamiliar with LDA-style outputs. For example, not all would agree that \textit{black, racial, white, community, individuals} should relate to passages that discuss race, but do not focus on Black or White identity. That is, sometimes only a few, but not all, words in an LDA-style topic label may be relevant to a passage. Our annotators tended to rate relevance based off of the most relevant word in a \texttt{Retell} topic's top five words.

\subsection{GPT-4o Performance and Behavior}\label{sec:appdx_gpt-4o}

\begin{table}[t]
\centering
\resizebox{0.8\columnwidth}{!}{
\begin{tabular}{lcc}
\toprule
 & \textbf{TopicGPT-lite} & \textbf{\texttt{Retell}-\textit{describe}} \\
\midrule
\ding{51} & 0.60 & 0.63 \\
\textbf{?} & 0.25 & 0.23 \\
\ding{55} & 0.15 & 0.13 \\
\midrule
Precision & 0.24 & 0.15 \\
Recall & 0.12 & 0.05 \\
\bottomrule
\end{tabular}
}
\caption{When used with a stronger LM (GPT-4o), TopicGPT-lite has higher pairwise precision and recall than \texttt{Retell}-\textit{describe}. Its topic relatedness ratings by Prolific crowdworkers (first three rows) also improve to become similar to \texttt{Retell}-\textit{describe}.}
\label{tbl:gpt_4o}
\end{table} 

Our main text primarily focuses on resource-efficient members of various model families. Here, we conduct a preliminary analysis of GPT-4o, primarily comparing \texttt{Retell}-\textit{describe} and TopicGPT-lite. Since GPT-4o-mini tends to perform better with \textit{describe} than \textit{summarize} (Table~\ref{tab:topic_relatedness_extended}), we choose the former verb for our GPT-4o experiments. For TopicGPT-lite, since GPT-4o does not devolve in the same manner as smaller LMs with multiple-topic generation, we use the prompt in Figure~\ref{fig:topicgpt_generation2} to generate topics (much like the $k=89$ version of GPT-4o-mini's TopicGPT-lite). For topic assignment, we use the prompt in Figure~\ref{fig:topicgpt_assignment}, matching our setup with all other LMs. This process produced $k=152$ topics, and so for ``fair'' comparison, we run GPT-4o's \texttt{Retell}-\textit{describe} with $k=152$ as well. Evaluation results for these two approaches can be found in Table~\ref{tbl:gpt_4o}. We find that TopicGPT-lite improves with a stronger LM, with higher precision and recall than \texttt{Retell} and similar crowdsourced topic relatedness ratings. 

\section{Case Study: ELA Books} \label{appdx:case_study_ann}

\textit{\textbf{\textcolor{red}{Content warning}}: the following section includes depictions of racial/ethnic violence and racist language.}

\subsection{Passage Sampling} 

As discussed in the main text, preliminary annotations revealed that identifying passages of interest that mention and/or discuss race required reading many passages before hitting a positive example, much like searching for needles in a haystack. Thus, to yield a sufficient number of positive examples within the months we spent with our annotators, we annotated passages sampled from all texts that contain at least one keyword. Possible keywords include the following: 
\begin{itemize}
    \item 109 common words or phrases corresponding to racial, ethnic, cultural, or nationality identifiers found in our ELA dataset of 396 books. We obtained these by running the named entity recognition pipeline from BookNLP \cite{bamman2021booknlp} and manually filtering through the top 500 most common nominal PERSONs. Unlike externally scraped dictionaries and wordlists, his process yields terms that are actually used within our dataset, such as ``Oriental'', ``Scotchman'', and ``Creole''. 
    \item 484 names of countries and capitals from GeoNames, a geographical database. 
    \item 1,948 demonyms,\footnote{\url{https://en.wiktionary.org/wiki/Category:en:Demonyms}} ethnonyms,\footnote{\url{https://en.wiktionary.org/wiki/Category:en:Ethnonyms}}, Native American tribe names,\footnote{\url{https://en.wiktionary.org/wiki/Category:en:Native_American_tribes}} and nationalities\footnote{\url{https://en.wiktionary.org/wiki/Category:en:Nationalities}} from Wiktionary. 
    \item 78 words that tend to co-occur within the same 250 token window with the following seed words: ``race'', ``racial'', ``racist'', ``racists'', ``racism'', ``immigrants'', ``immigrant'', ``ethnic'', ``minority'', ``minorities'', ``diversity'', ``discrimination'', ``identity'', ``blackness'', ``whiteness''. Co-occurrence is measured by calculating the normalized pointwise mutual information between a potential keyword and one of these seed words in ELA books, and we use a threshold of 0.15 for determining significant co-occurrence. From this resulting list, we manually filter for words that may be suggestive of race-related discussion, e.g. ``segregation'', ``migrant'', and ``supremacy.''
\end{itemize}

\subsection{Annotation Codebook} \label{sec:appdx_cs_ann}

\subsubsection{Defining ``race''}

After sampling passages to annotate, the second author, a social psychologist, constructed an annotation codebook for determining what constitutes explicit mentions and discussions of race. Race is a socially constructed, contested, and multidimensional concept \cite{field-etal-2021-survey, hanna2020race}. So, our instructions for annotators included several definitions of race, presenting it as a phenomenon that overlaps with ethnicity, religion, geography, biology, and culture. We generally follow precedent set by \citet{algee2020representing}, who conducted a cultural analytics study of racial and ethnic representation in American fiction:

\begin{displayquote}
The categories include broad racial/ethnic groups (e.g., black, white, Native American, Pacific Islander, Latinx), ethno-religious distinctions (e.g., Catholic, Jewish, Middle Eastern and Muslim), geographical origins (e.g., South Asian, Filipino, East Asian, Eastern European, German/Dutch, Irish, Italian, Latin American, Scandinavian), and one catch-all field (Immigrants).
\end{displayquote}

To \citet{algee2020representing}'s definition above, we added racial/ethnic groups of Pacific Islanders and Hispanic/Latinx people. We also theoretically grounded ourselves in definitions of race from sociologists and educators, including: 
\begin{itemize}
    \item ``Race is a biological or cultural category easy to read through marks in the body (phenotype) or the cultural practices of groups'' \cite{bonilla2015}.
    \item ``Race is a system for classifying human beings that is grounded in the belief that they embody inherited and fixed biological characteristics that identify them as members of racial groups'' \cite{morning2011}.
    \item ``Culture can be defined as deep-rooted values, beliefs, languages, customs, and norms shared among a group of people'' \cite{milner2017}. 
\end{itemize}

We want to be clear that we do not conflate the categories of race, ethnicity, and culture, and we understand them as distinct socially constructed identifiers \citep[e.g.][]{spencer2014race, worrell2014culture}. Our coding effort required us to operationalize these dimensions and aggregate across them, so as not to produce too many themes.

\subsubsection{Codebook development and codes}\label{sec:appdx_codebook}

We used a directed content analysis approach \cite{hsieh2005three} in developing our coding scheme. We start with face-valid and theory-driven themes and using the training phases of the annotation process to fine-tune, iterate, and further define these initial themes, and to add new themes to capture instances that could not be coded by the initial themes. We only code for explicit mentions or discussions of race, ethnicity, and culture, due to the fact that more implicit instances would sometimes require the coder to use information outside of the passage itself to infer that there was a reference to race, ethnicity, or culture. In this section, we include a few examples of what types of content would or would not fall within each code; we provided more extensive examples to our annotators during the coding process. Passages can be labeled with multiple codes. 

We started with three themes that captured explicit and face-valid instances of race, ethnicity, or culture being described or discussed. In each case, we wanted to capture language where there was an explicit engagement with identity: 

\paragraph{A. Descriptions of racial/ethnic/cultural identity groups.} This code includes descriptions of a racial/ethnic group, or a community as such. It does not include descriptions of an individual person, unless that person is explicitly racialized and framed as representative of their group. These descriptions can be in a first person or third person perspective, and include descriptions of a trait, experience, practice, tradition, or condition shared by a group. 
\begin{itemize}
    \item Affirmative example: \textit{... he didn't know that all Indian families are unhappy for the same exact reasons: the fricking booze} in \textit{The Absolutely True Diary of a Part-Time Indian} by Sherman Alexie. Note that this example would also be coded as \textbf{C}, as defined below. 
    \item Non-example: \textit{Nikki was lightskinned and from Texas, about five-foot-nine} in \textit{No Disrespect} by Sister Souljah.
\end{itemize}
\paragraph{B. Instances of racially-based violence.} This code includes the discussion or mention of racially-based violence. 
\begin{itemize}
    \item Affirmative example: \textit{... let those [n-word] stand there with guns, and we don't accommodate them? They want war, let's give them war} in \textit{A Gathering of Old Men} by Ernest J. Gaines. 
    \item Non-example (too implicit): \textit{... true that some of the elders of the Somali community remained angry ... } in \textit{The Burgess Boys} by Elizabeth Strout. 
\end{itemize}

\paragraph{C. Negative racist language or dialogue.} This code includes racist or prejudiced language spoken by, to, or about a character. It can involve non-dialogical language indicating racist beliefs or behaviors of characters. The language may be directed towards a group or towards a person. 
\begin{itemize}
    \item Affirmative example: \textit{We're not such paupers as to sell to Japs, are we?} in \textit{Snow Falling on Cedars} by David Guterson. 
\end{itemize}

During the early training phase of coding, we encouraged coders to suggest additional themes for instances where a passage did not fit within an existing theme. From this process, we added two additional themes:

\paragraph{D. Mentions of racial, ethnic, or cultural identity.} This code includes cases where race or racial identity appears in a passage, but may not reach the level of engagement as initial codes. The group should be explicitly mentioned as group identity - i.e., talked about as a racial, ethnic, or cultural group and not simply as a descriptor or mention of a country. Character names or language can be used as cues to group identity only if they corroborate other cues to group identity. Can include mentions of racial/ethnic groups in fictional or inanimate contexts (e.g., on TV). 
\begin{itemize}
    \item Affirmative example: \textit{... who grew up in such poverty that other poor Indians called her family poor?} in \textit{Reservation Blues} by Sherman Alexie. 
    \item Non-example: \textit{I sat dumbfounded among the Chinese antiquities and Greek vases...} in \textit{The Goldfinch} by Donna Tartt. 
\end{itemize}

\paragraph{E. Mentions of racism or inequality.} Explicit or blatant mention or labeling of something as racism, anti-racist, inequality, racial injustice, racial segregation, desegregation, or racial privilege, without the characters necessarily engaging in racist language or acts. 
\begin{itemize}
    \item \textit{... we couldn't have racism in America without racists} in \textit{All American Boys} by Jason Reynolds and Brendan Kiely.
\end{itemize}

In the main text (\S\ref{sec:case_study}), passages labeled as codes \textbf{A}, \textbf{B}, \textbf{C}, and \textbf{E} constitute passage set \texttt{discuss}, or deeper engagement with topics around racial/cultural identity, while passages labeled as \textbf{D} but \textit{not} \textbf{A}, \textbf{B}, \textbf{C}, nor \textbf{E} constitute passage set \texttt{mention}. Passages that do not fall under any of the codes are considered negative examples in \texttt{neither}.

\subsection{Annotation Process} 

\begin{table}[t]
\centering
\resizebox{\columnwidth}{!}{
\begin{tabular}{llcc}
\toprule
 & \textbf{Code} & \textbf{\% agreement} & \textbf{Cohen's $\kappa$} \\
\midrule
\multirow{5}{*}{\textbf{Stage 1}} & A. Race description & 97.1 & 0.603 \\
& B. Race-based violence & 98.9 & 0.794 \\
& C. Racism language & 99.4 & 0.854 \\
& D. Race mention & 94.3 & 0.865 \\
& E. Racism mention & 97.7 & 0.870 \\
\midrule
\multirow{5}{*}{\textbf{Stage 2}} & A. Race description &  93.0 &  0.500 \\
&  B. Race-based violence&  99.0 &   NA \\
& C. Racism language &  97.0 &  NA \\
& D. Race mention & 93.0 & 0.855 \\
& E. Racism mention &  97.7 &  0.740 \\
\bottomrule
\end{tabular}
}
\caption{Inter-annotator agreement during the two annotation stages for our case study of race in ELA literary passages. We report the \% of agreed-upon passages and unweighted Cohen's $\kappa$.}
\label{tbl:iaa}
\end{table} 

We annotated passages in two stages, as our annotation process spanned two academic terms, and different student annotators were available in each term. 

\paragraph{Stage 1.} Two undergraduate students with academic backgrounds in psychology and cognitive science annotated 711 passages in the first term. This time period was focused on refining and finalizing the coding scheme. We collectively coded some passages, refined descriptions and definitions for existing themes, and added new themes when necessary (Appendix~\ref{sec:appdx_codebook}). Our process matched how directed content analysis is generally conducted \cite{hsieh2005three}. 

Annotators were assigned batches of passages to annotate and discuss with the first and second authors in a weekly meeting, with the number of passages increasing with every week. We started with 10 passages the first week and 150 passages by the 6th week. Both annotators coded the same 711 passages. In each team meeting, the second author compiled each instance where annotators disagreed about the presence or absence of a given code for the same passage. For each disagreement, the annotators discussed their reasoning for why they did or did not include a code, came to an understanding about the `correct' code, and recorded the agreed-upon code. In the instance where there wasn't an existing theme for a given passage, the team considered additions to the coding scheme. 
When new themes were added, annotators went back to previously coded passages to determine whether any passages should be assigned the newly defined themes. Table~\ref{tbl:iaa} includes levels of inter-annotator agreement we obtained at the end of five weeks.

\paragraph{Stage 2.} We recruited two new undergraduate student annotators, whose academic backgrounds also included cognitive science and psychology. During this stage, we focused solely on achieving agreement between new annotators and annotating a larger sample of passages. Our collective annotation process mirrored that of Stage 1, and during this phase, we annotated 435 passages. Table~\ref{tbl:iaa} details our inter-annotator agreement after six weeks of collective annotation. We note that percent agreement calculations do not take into account the number of instances, while Cohen's $\kappa$ does, and so our agreement for \textbf{A} has a lower $\kappa$ score than other categories. In addition, there were not enough instances of \textbf{B} and \textbf{C} to properly calculate Cohen's $\kappa$, hence those values are listed as `NA'. We concluded our second annotation stage with the students independently coding 500 passages, or 250 per student. 

%%%%%%%%%% PROMPTS %%%%%%%%%%%%%%%

\begin{figure*}[h]
    \centering % Center the textbox within the figure environment
    \begin{tcolorbox}[
    prompt,
    title={\textbf{TopicGPT-lite: generating one topic per passage}},
    ]
    \small

You will receive a document and a set of topics. Your task is to identify a generalizable topic within the document. If this topic is missing from the provided set, please add it. Otherwise, output an existing topic as identified in the document. \\

[Topics] \\
{Topics} \\

[Examples] \\
Example 1: Adding "Religion" \\
Document:  \\
But a religion true to its nature must also be concerned about man’s social conditions... Such a religion is the kind the Marxists like to see—an opiate of the people.  \\

Your response:  \\
Religion: Describes purposes and roles of religion for people.  \\

Example 2: Respond with an existing topic, "Love"  \\
Document:  \\
``I have reason to think,” he replied, “that Harriet Smith ... is he sure that Harriet means to marry him?'' \\

Your response:  \\
Love: Discusses romantic relationships, such as marriage. \\

[Instructions] \\
Step 1: Determine the topic mentioned in the document.  \\
- The topic label must be as GENERALIZABLE as possible. It must not be document-specific. \\
- The topic must reflect a SINGLE topic instead of a combination of topics. \\
- A new topic must have a short general label and a topic description.   \\
Step 2: Perform ONE of the following operations:  \\
1. If there are already a duplicate or relevant topic in the provided set of topics, output that topic and stop here.  \\
2. If the document contains no topic, return "None".  \\
3. Otherwise, add a new topic to the set of topics.  \\

[Document] \\
{Document} \\

Your response should be one line and ONLY contain a topic, written in the format "[Topic label]: [your reasoning]".  \\
Your response: \\

    \end{tcolorbox}
    \caption{Prompt for generating a possible topic pool over 1k passages. We truncate the few-shot examples here for brevity (with ``...''), but show the beginnings and ends to support their reconstruction. The first example is from Martin Luther King Jr.'s \textit{Stride Toward Freedom}, while the second is from Jane Austen's \textit{Emma}.}
    \label{fig:topicgpt_generation}
\end{figure*}

\begin{figure*}[h]
    \centering % Center the textbox within the figure environment
    \begin{tcolorbox}[
    prompt,
    title={\textbf{TopicGPT-lite: generating multiple topics per passage}},
    ]
    \small

You will receive a document and a set of topics. Your task is to identify generalizable topics within the document. If any relevant topics are missing from the provided set, please add them. Otherwise, output any existing topics identified in the document. \\

[Topics] \\
{Topics} \\

[Examples] \\
Example 1: Adding "Religion" \\
Document:  \\
But a religion true to its nature must also be concerned about man’s social conditions... Such a religion is the kind the Marxists like to see—an opiate of the people.  \\

Your response:  \\
Religion: Describes purposes and roles of religion for people.  \\

Example 2: Respond with an existing topic, "Love"  \\
Document:  \\
``I have reason to think,” he replied, “that Harriet Smith ... is he sure that Harriet means to marry him?'' \\

Your response:  \\
Love: Discusses romantic relationships, such as marriage. \\

[Instructions] \\
Step 1: Determine topics mentioned in the document.  \\
- The topic labels must be as GENERALIZABLE as possible. They must not be document-specific. \\
- The topics must reflect a SINGLE topic instead of a combination of topics. \\
- Each new topic must have a short general label and a topic description.   \\
Step 2: Perform ONE of the following operations:  \\
1. If there are already duplicates or relevant topics in the provided set of topics, output those topics and stop here.  \\
2. If the document contains no topic, return "None".  \\
3. Otherwise, output new, additional topic(s). \\

[Document] \\
{Document} \\

Your response should ONLY contain topics, written in the format "[Topic label]: [your reasoning]".  \\
Your response: \\

    \end{tcolorbox}
    \caption{A modified version of Figure~\ref{fig:topicgpt_generation} for generating a possible topic pool over 1k passages, where instead of asking an LM to generate one topic per passage, it can generate multiple.}
    \label{fig:topicgpt_generation2}
\end{figure*}

\begin{figure*}[h]
    \centering % Center the textbox within the figure environment
    \begin{tcolorbox}[
    prompt,
    title={\textbf{TopicGPT-lite: assigning topics to passages}},
    ]
    \small

You will receive a document and a set of topics. Assign the document to the most relevant topics. Then, output the topic labels and assignment reasoning. DO NOT make up new topics. \\

[Topics] \\
{tree} \\

[Examples] \\
Example 1: Assign "Religion" to the document \\
Document:  \\
The second step for me to morph from animal to human was taking me to the church youth group ... pray in separate rooms so religions wouldn't collide. \\

Assignment: \\
Religion: Describes people's religious practices. \\

Example 2: Assign "Love" to the document \\
Document:  \\
``Good-bye, my love,'' answered the countess ... smile fluttering between her lips and her eyes, she gave her hand to Vronsky.  \\

Assignment:  \\
Love: Describes a scene where one person declares and shows affection for another. \\

[Instructions] \\
1. Topic labels must be present in the provided set of topics. You MUST NOT make up new topics.  \\
2. Output topic(s) you assign in order of their prominence in the document, with the most prominent topic first. \\
3. Each line of your response should contain a topic written in the format "[Topic label]: [your reasoning]".  \\

[Document] \\
{Document} \\

Your response should ONLY contain topics. Double check that your assignment exists in the provided set of topics! \\
Your response: \\

    \end{tcolorbox}
    \caption{Prompt for assigning topics to passages. Few-shot examples are shortened with ``...'' for brevity purposes, and the first is from \textit{Gabi, a Girl in Pieces} by Isabel Quintero, while the second is from \textit{Anna Karenina} by Leo Tolstoy.}
    \label{fig:topicgpt_assignment}
\end{figure*}

\end{document}